\documentclass[11pt,a4paper]{article}

\usepackage{times,latexsym} \usepackage{url} \usepackage[T1]{fontenc}

\usepackage{natbib} \usepackage{amsmath} \usepackage{amssymb}
\usepackage{amsthm} \usepackage{multicol}

\usepackage[pdftex]{graphicx}
\usepackage{color}

\usepackage[hang,small,bf]{caption}
\usepackage[subrefformat=parens]{subcaption}
\usepackage{comment}

\newcommand{\secref}[1]{Section \ref{#1}}
\newcommand{\appref}[1]{Appendix \ref{#1}}
\newcommand{\figref}[1]{Figure \ref{#1}} \newcommand{\tabref}[1]{Table
  \ref{#1}}

\newcommand{\red}[1]{\textcolor{black}{#1}}
\newcommand{\blue}[1]{\textcolor{blue}{}}

\definecolor{Green}{rgb}{0.0, 0.3, 0.0}

\definecolor{Yellow}{rgb}{0.8, 0.8, 0.0}

\usepackage{vwcol}

\date{}

\title{Strahler Number of Natural Language Sentences in Comparison with Random Trees}

\providecommand{\keywords}[1]{\textbf{\textit{Keywords---}} #1}

\author{Kumiko Tanaka-Ishii and Akira Tanaka \\
Waseda University \hspace*{0.5cm} University of Tokyo
}
\begin{document}

\maketitle

\begin{abstract}
The Strahler number was originally proposed to characterize the
complexity of river bifurcation and has found various
applications. This article proposes computation of the Strahler
number's upper and lower limits for natural language sentence tree
structures. Through empirical measurements across grammatically
annotated data, the Strahler number of natural language sentences is
shown to be almost 3 or 4, similarly to the case of river bifurcation
as reported by Strahler (1957). From the theory
behind the number, we show that it is one kind of lower limit on the
amount of memory required to process sentences. We consider the
Strahler number to provide reasoning that explains reports showing
that the number of required memory areas to process sentences is 3 to
4 for parsing
\blue{\citep{abney1991memory}}\citep{schuler2010broad},
and reports
indicating a psychological ``magical number'' of 3 to 5
\citep{cowan01}. An analytical and empirical analysis shows that the
Strahler number is not constant but grows logarithmically; therefore,
the Strahler number of sentences derives from the range of sentence
lengths. Furthermore, the Strahler number is not different for random
trees, which could suggest that its origin is not specific to natural
language.
\end{abstract}
\keywords{ Strahler number; Tree structure; Memory}

\section{Introduction}
The Strahler number \citep{Strahler57} was originally introduced in
the field of geography, as a measure of the complexity of river
bifurcation. Curiously, Strahler found that almost any river in
England has a constant value of 4 for this number. Apart from
geography, the Strahler number has been applied to analyze the
complexity of computation trees in computer program source code
\citep{Ershov58}. In particular, it was theorized to equal the minimum
number of memory areas that are necessary for evaluation of a
computation tree \citep{Ershov58}.

We believe that application of the Strahler number to natural language
sentences can contribute, first of all, to understanding the Strahler
number itself. Although the Strahler number has been reported to yield
a value of four, various questions about it have not been fully
answered: for example, what does this constant value signify apart
from its definition, how does it grow with respect to the tree size,
and what is the relation between random and real trees? In this work,
we show that this number grows logarithmically with respect to the
tree size. Furthermore, we empirically demonstrate that the Strahler
number for natural, real trees is not different from that for random
trees. In other words, the trees' size range decides the Strahler
number, which does not differ for random trees of the same size.

We believe that this understanding is important to natural language
sentence structure, because multiple studies in psychology and natural
language processing have empirically reported almost the same number,
but without any theoretical grounding. Previous works on the
structural characteristics of natural language sentences have focused
on the cognitive load \citep{yngve60, Kimball75, gibson00,
  liu2017dependency}. \cite{cowan01} suggested a value of 3 to 5 for a
``magical number'' involved in cognitive short-term
memory. Furthermore, in natural language processing of parsing
methods, \blue{\cite{abney1991memory} and}
\cite{schuler2010broad} indicated
that human sentences require a maximum of four memory areas for a
particular sentence. However, these previous works did not state why
the number is four.

As will be shown here theoretically, the Strahler number of human
sentences shows a kind of lower limit on the amount of memory
necessary to understand sentence structure under a certain setting. We
provide a mathematical definition of this lower limit and show that it
is {\em not} actually a constant, but rather, it increases
logarithmically with the sentence size. It is a fact, however, that
sentence lengths can only take a certain range \citep{sichel,yule},
and this range is one factor in why the Strahler number is seemingly a
constant. Furthermore, our work shows that this number is almost the
same for random trees, thus providing a signification that the
potential ``magical number'' might not be so ``magical,'' by
explaining its origin via random trees.

\section{Related Work}
This work is related to four fields as follows. The first involves the
general history of the Strahler number \citep{Strahler57}. It was
known in the literature before Strahler \citep{horten45}; however, we
call it the ``Strahler number'' following convention. The Strahler
number was analyzed from a statistical viewpoint in relation to the
bifurcation ratio and area of a water field \citep{beer93}. Meanwhile,
it has found various applications besides river morphology, of which
the most important is computer trees \citep{Ershov58}, as mentioned
above. That theory is the basis of this article, as explained in the
following section.

The second genre of related work is measurement to characterize the
complexity of natural language. Previously, there have been diverse
approaches to consider this complexity. The primary approach is the
Shannon entropy, which has seen numerous
applications. \cite{entropy16} provided a summary, and the Shannon
entropy was recently applied to characterize the specific field of
legal texts \citep{friedrich21}. Apart from the Shannon entropy,
various methods from statistical physics have also been applied to
texts, including Zipf's law \citep{zipf}, long memory via methods such
as long-range correlation \citep{altmann2009,Altmann2012,plosone16},
and fluctuation analysis
\citep{ebeling1994,ebeling1995,acl18,jpc18}. All of those works
examined sequences, however, rather than sentence structure.

The study of sentence structure relates to the long memory phenomena
of natural language. This is because one factor in long memory is the
tree structure underlying a sequence: \cite{lin_2016} analytically
showed how a context-free grammar would produce long memory. More
recent works have conjectured the nature of a formal grammar
\citep{degiuli19-1,degiuli19-2}, including whether it has phase
changes. However, understanding of what kind of actual tree structure
is the underlying cause of what kind of long memory will require more
direct study of trees derived from real data. We believe that datasets
of grammatically annotated sentence structures provide a good starting
point.

For natural language sentence structure, there is a known bias in the
branching direction in sentences, such as a right-branching preference
in Indo-European (IE) languages
\citep{forster1968sentence}. \blue{This} \red{Such} bias has been
quantified in various ways, as excellently summarized in
\cite{fischer21}. One way to consider the complexity of this
phenomenon is via the modifier-modified distances within a sentence
\citep{gibson00}. Through an analysis of 20 languages,
\citet{liu2008dependency} reported that the dependency distance
remains short but never reaches its theoretical minimum. More recent
works have shown syntactic complexity \blue{based on the dependency
  distance} \red{ based on various statistics acquired from sentence
  structure} \citep{xu2016convergence,yadav20}. The Menzerath law
provides another quantitative approach to analyze the complexity of
sentence structure. Specifically, the Menzerath conjecture of ``the
greater the whole, the smaller its parts'' has been considered as a
relation between the number and size of sentences' main constituents
in various languages \citep{mal21,hou17,macutek17,sanada,buk07}. In
contrast, our work on the Strahler number takes a different approach
from both of those quantitative methods applied to sentence structure.

The third genre of related work involves the amount of short-term
memory as studied in the field of cognitive science. Among early
works, \citet{miller56} showed that the number of {\em chunks in
  short-term memory} is 7 $\pm$ 2. \citet{yngve60} defined the
complexity of dependency trees by their depth and argued that this
depth is related to Miller's number. Beyond language, \citet{cowan01}
argued that short-term memory is bounded by a ``magical number'' of 3
to 5. The exact nature of this {\em short-term memory} has been
controversial. Our work provides a novel approach
to explain this memory by using the
Strahler number and its mathematical theory for random trees.

Lastly, this work relates to a genre of works in natural language
processing on parsing. Every parsing study presents a method to
reverse engineer a sentence structure. By applying such methods, some
works showed the maximum amount of memory necessary for parsing. In
particular, \blue{\cite{abney1991memory},} \cite{schuler2010broad}, and more
recently \cite{noji-miyao-2014-left} indicated that \red{processing} human sentences
require almost a maximum of three to four memory areas. This number \blue{of four} has only
been an empirical result, with no theoretical grounding. Hence, this
work intends to explain this empirical result.

\section{Strahler Number}
\subsection{Definition}
\label{sec:definition}
\label{sec:Strahler}
Let $t = (V,E)$ denote a rooted directed tree, where $V$ is the set of
nodes and $E \in V \times V$ is the set of edges. Each edge is
directed from a parent to a child. Let $T$ denote the set of finite
rooted directed trees, and let $n$ denote the number of nodes in a
tree. Later, we will consider different sets as $T$: (1) dependency
structures $U$ (\secref{sec:ud}), with $U(n)$ denoting the subset with
$n$ nodes; (2) random binary structures $R_2(n)$
(\secref{sec:random}); and (3) random $n$-node trees $R(n)$
(\secref{sec:nrandom}), as defined later.

Let a {\em binary} tree be one for which every inner node has two
children. For a binary tree $t$, the Strahler number is defined in a
bottom-up manner \citep{Strahler57,horten45}. Every node $v$ acquires
a Strahler number $S(v)$, and the Strahler number of the root is the
Strahler number of the whole tree, $S(t)$. The definition is given as
follows:
\begin{quote}
\begin{itemize}
\item For a leaf node $v$, $S(v)=1$.
\item For an inner node $v$, let the two child nodes be $ch_1(v),
  ch_2(v)$.
 \begin{itemize}
 \item If $S(ch_1(v)) == S(ch_2(v))$, then $S(v) = S(ch_1(v)) + 1$.
 \item Otherwise, $S(v) = \text{max}(ch_1(v), ch_2(v)).$
 \end{itemize}
 \end{itemize}
\end{quote}
From this definition, the Strahler number is obviously unique for a
given tree.

\begin{figure}[t]
\centering \includegraphics[width =
  0.4\linewidth]{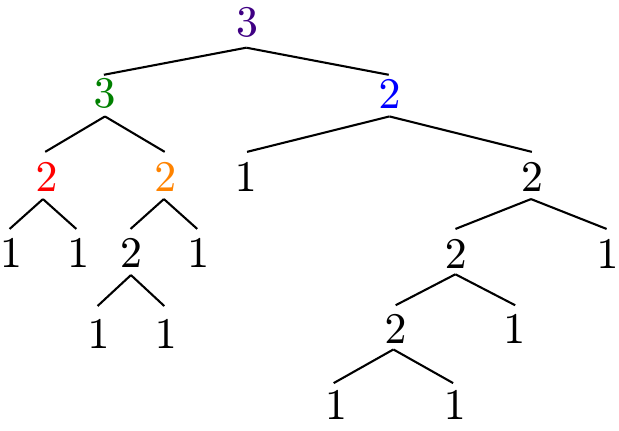}
\caption{Strahler number of a tree.}
\label{fig:strahler}
\end{figure}

\figref{fig:strahler} shows an example tree with the values of $S(v)$
indicated for every node $v$. For instance, the node with the green
``3'' has two children. As the child nodes' numbers are both 2, the
parent node's number is $2+1=3$. On the other hand, the root node with
the purple number also has two children, one with a number of 3
(green) and the other with 2 (blue). Because the child nodes' numbers
are different, the Strahler number of the root is $\max(3, 2) =
3$. Through such bottom-up calculation, this tree's Strahler number is
calculated as 3.

\subsection{Relation to Number of Memory Areas Required to Process Trees}
After the Strahler number's original definition to analyze river
bifurcation in England, it was applied to analyze the complexity of
computation trees \citep{Ershov58}. A computation tree is produced
from program code, which is parsed into a computation tree and then
evaluated.

\begin{figure}[t]
\begin{minipage}[t]{0.25\linewidth}
\centering \includegraphics[bb=0 0 150 160,
  scale=1.0]{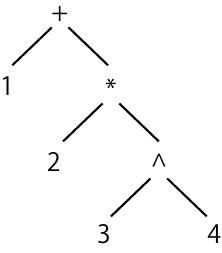}
\caption{Computation tree for $1+2*3^4$. \label{fig:ct}}
\end{minipage}
\hspace{0.01\linewidth}
\begin{minipage}[t]{0.35\linewidth}
\centering \includegraphics[bb=330 0 160 180,
  scale=0.33]{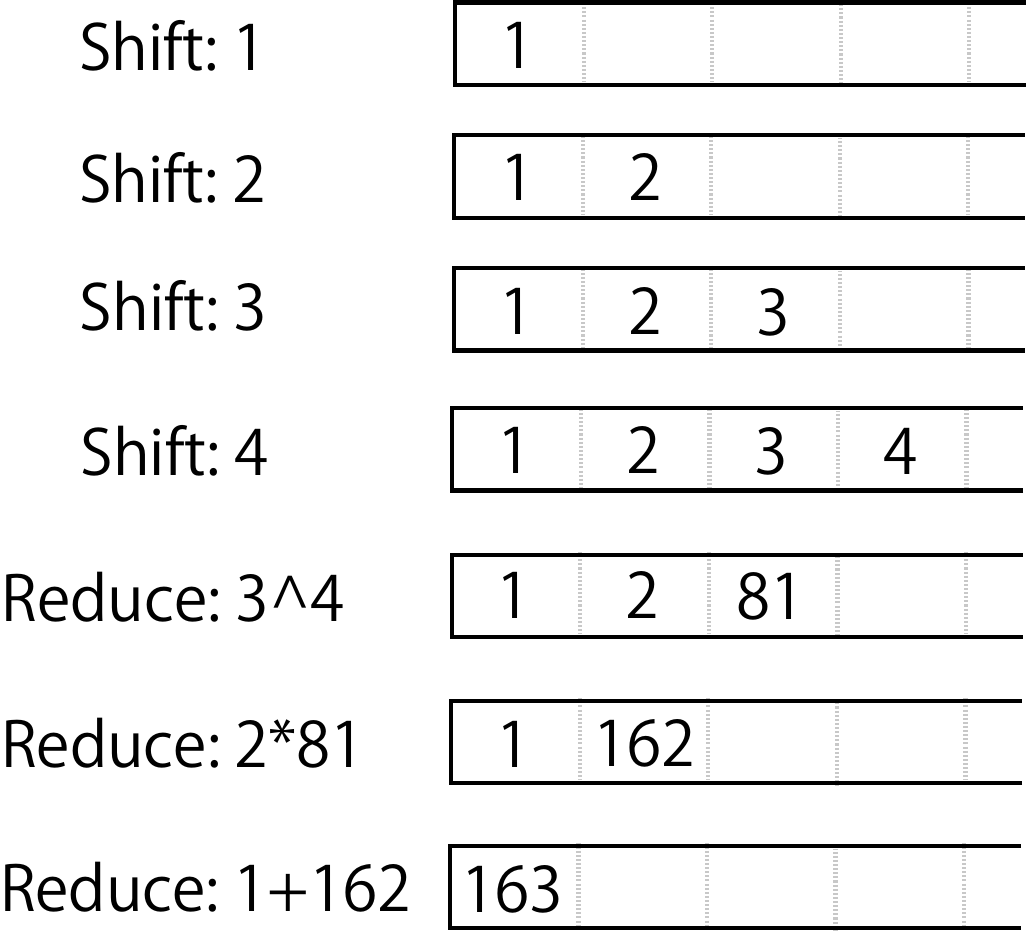}
\caption{Transition of the stack during evaluation from the beginning
  of the tree. \label{fig:ct-eval1}}
\end{minipage}
\hspace{0.01\linewidth}
\begin{minipage}[t]{0.35\linewidth}
\centering \includegraphics[bb=330 0 160 180,
  scale=0.33]{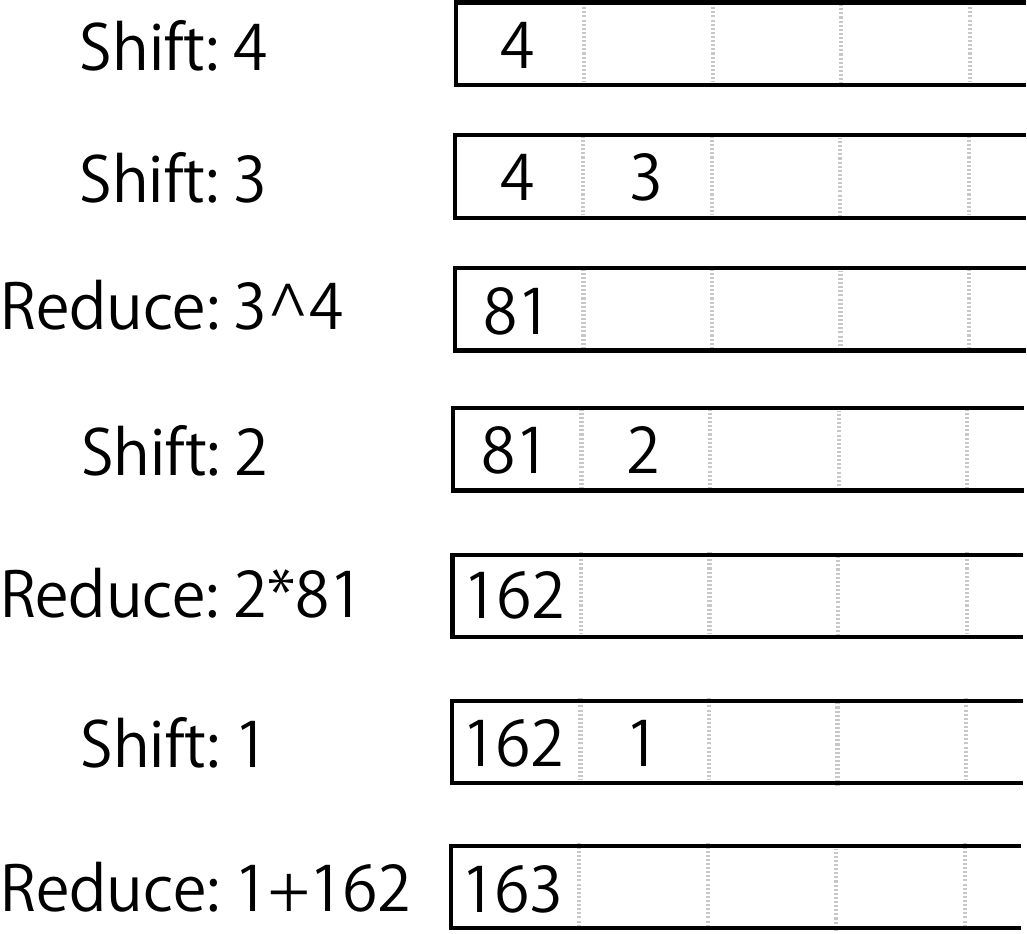}
\caption{Transition of the stack during evaluation from the
  end. \label{fig:ct-eval2}}
\end{minipage}
\end{figure}

For example, \figref{fig:ct} shows a tree for a computation (i.e.,
program code) ``$1 + 2 * 3^4$''. Parsing this program string generates
the tree, which is then computed to yield 163. The question here is
how much memory is necessary to get this result.

The Strahler number is known to give the minimum number of memory
areas for tree evaluation by the use of {\em shift-reduce} operations
\citep{sethi70}, which constitute the simplest, most basic theory of
computation tree evaluation. Here, we give a brief summary of these
operations, with a more formal introduction given in
\appref{sec:relations}. A computation tree can be evaluated with the
two operations of shift and reduce by using a memory system comprising
a stack, which is a last-in, first-out (LIFO) data structure. A {\em
  shift} operation puts the data element of a tree leaf on the stack,
and a {\em reduce} operation applies a functional operation (such as
addition or multiplication) to the two elements at the top of the
stack.

For example, consider evaluation of the computation tree shown in
\figref{fig:ct}. For evaluation {\em from the beginning of the tree},
the required number of stack spaces is four, as shown in
\figref{fig:ct-eval1}. On the other hand, for evaluation from the end
of the tree, the total number is reduced to two, as shown in
\figref{fig:ct-eval2}.

As seen here, which leaf of the tree is evaluated first determines the
necessary depth of the stack. Every shift-reduce gives a way to
traverse a given computation tree, and each way requires a particular
number of stack space uses. Thus, there is a particular way to
traverse a tree by the shift-reduce method that requires a minimum
number of stack spaces.

This minimum number of stack spaces required to evaluate a computation
tree equals the tree's Strahler number \citep{Ershov58}, which is
obvious from the definition of the shift-reduce method as given in
\appref{sec:relations}. If no self-referential expression is involved,
then this number is also the minimum number required for analyzing
program code in a sequence represented as a computation tree. This is
because analysis of a program as a computation tree is yet another way
to traverse the tree.

To adapt this theory to natural language sentences, we can consider
transformation of a sentence structure into a binary tree. The
evaluation of this binary tree (to obtain some kind of meaningful
representation) uses a certain memory amount. In describing this
amount with use of the shift-reduce method, the necessary number of
stack spaces for evaluation is bounded by the Strahler number. Because
analyzing a sentence is equivalent to traversing a binary tree, the
tree's Strahler number gives the lower bound on the necessary number
of stack spaces. This shift-reduce scheme is the simplest general
method to deal with a sentence structure \citep{zhang2020survey}. It
has become a standard way to parse a sentence, and its use is an
ongoing research topic \citep{fernandez2019faster, yang2020strongly,
  grenander2022sentence, fernandez2023discontinuous}. Hence, knowledge
of a sentence structure's Strahler number can give a lower-limit
criterion for the amount of memory required to process the sentence
structure.

\subsection{Strahler Number of Random Binary Trees with $n$ Leaves: $R_2(n)$}
\label{sec:random}
Before calculating the Strahler number of a sentence structure, we
introduce the Strahler number of a random binary tree, which provides
a good theoretical baseline.

Let $R_2(n)$ be the set of all binary trees with $n$ leaves. The set's
size $|R_2(n)|$ is known to be given by a Catalan number, i.e.,
$|R_2(n)|= \frac{1}{n}{}_{2n-2}C_{n-1}$ \citep{stanley2015catalan}.

\cite{flajolet79} analytically showed that the mean Strahler number
can be deductively described via approximately logarithmic growth with
a base of four\footnote{ Precisely, \cite{flajolet79} showed that the
mean Strahler number is \begin{equation} E[R_2(n)] = \log_4 n + 1 -
  \int_{0}^{\infty} (e^{-t^2} H_4(t))( tF(\log{t}+ \frac{1}{2}\log{n}
  ) + \frac{t}{2}\log{t})\textrm{d}t+\textrm{o}(1), \end{equation}
where $F$ is a continuous, periodic function having period 1, and
$H_4$ is the fourth Hermite polynomial.}, and the mean value obviously
increases with the tree size $n$. Later, this theoretical fact will
provide an important reference in understanding the complexity of
natural language sentences.

In addition to the Strahler number's mean behavior, its upper and
lower limits can be considered. For a given set of trees, $T$, the
upper/lower limits are respectively defined as the maximum/minimum
Strahler numbers. Hence, we analytically consider the upper/lower
limits for $R_2(n)$. By the definition of the Strahler number, the
upper limit is obviously acquired from a tree that is closest to a
complete tree \citep{ehrenfeucht1981etol}, where the Strahler number
equals the tree's maximum depth. Therefore, the upper limit for
$R_2(n)$ is $\lfloor \log_2{n}\rfloor+1 $. On the other hand, the
lower limit derives from the opposite case of a tree that is closest
to a {\em linear} tree. Specifically, the lower limit is 1 for $n=1$,
or 2 otherwise, because for $n>1$, there are two leaf nodes and all
inner nodes thus have a Strahler number of 2.

\label{sec:maxminT2}

\section{Measurement of Strahler Number of Sentence Structure}
\label{sec:nl}
There have been two main paradigms in representing tree structures:
phrase structure \citep{Chomsky56} and dependency structure
\citep{Tesniere59}. Here, we use these terms under the most
conventional definitions, but briefly, the former describes natural
language sentences in a similar manner to a computation tree, as
described above, where words are located at leaves and inner nodes
describe the relations between words. On the other hand, the latter
describes a tree structure as the modifier-modified relations among
words. In other words, the inner nodes of the tree in the phrase
structure paradigm are not words, whereas those in the dependency
structure paradigm are words.

In this article, we calculate the Strahler number with a dependency
structure rather than a phrase structure, because a large amount of
annotated data is available in a large number of languages, as with
the data that will be described in \secref{sec:ud}. Hence, the
question here is how to calculate the Strahler number for every
dependency tree.

In \secref{sec:definition}, the Strahler number was defined for a
binary tree, whose inner nodes and leaf nodes are different, with only
leaf nodes representing words. On the other hand, both the inner and
leaf nodes of a dependency structure are words, with inner nodes
having multiple child nodes for modifiers. Filling of the gap between
the differences in these two settings would suggest only two
directions: to transform dependency structures into binary phrase
structure trees; or to extend the Strahler number by adapting it to
the dependency structure.

Regarding the latter direction, there have been previous attempts to
extend the Strahler number to general trees with nodes having more
than two children \citep{auber04}. The method in that work extended
the rule to count up the Strahler number at each bifurcation as
described in \secref{sec:Strahler}. However, we do not adopt this
generalization, mainly because the theory around it is not
established. The theory on computation trees would not apply easily;
in addition, the analytical theory for random binary trees would not
be easy to extend to general trees.

Hence, in the following, we consider methods to transform dependency
trees into binary trees to calculate the Strahler number. First, we
explain two particular binarization methods. Later, in the
experimental section (\secref{sec:exp}), we show that these two
methods yield very similar results with respect to the Strahler
number. Second, we provide a method to acquire the upper and lower
limits {\em across} any binarization method. The results for any
particular binarization method fall within the range between the upper
and lower limits, and the limits can be compared with those of random
trees.

\subsection{Two Binarization Methods for Dependency Structure}
\label{sec:binary_structure}
The transformation of a dependency structure to a phrase structure is
{\em not} easy \citep{n15-1080,p15-1147}, partly because the
grammatical attribute of every inner node must be estimated, whereas
the reverse transformation is relatively feasible
\citep{buchholz02}. Here, we want to effectuate this difficult
transformation but without requiring any precise prediction of the
attributes of inner nodes, as we want to calculate the Strahler number
gregardless of its specific value.

We transform a given dependency structure via the following two
methods:
\begin{itemize}
\item[Binary1] Transformation by use of a manually crafted grammar
  \citep{tran22}.
\item[Binary2] Transformation without a grammar, by use of heuristics.
\end{itemize}

Binary1 derives from a grammar proposed by \citet{reddy17}. The
grammar describes the degree of grammatical relation between the
modifier and modified, and the dependency tree is binarized on the
order of this degree. For an explanation of this grammar, see
\cite{reddy17}.

On the other hand, Binary2 binarizes a dependency structure via two
simple heuristics based on a modifier's distance from the head. The
two heuristics are as follows: \blue{(1) the farthest modifiers form deeper
nodes in the tree; and (2) words before the head are allocated to the
head's left, whereas those after are allocated to the right.}
\red{(1) the modifiers before the head are bifurcated then those after; 
(2) the farther modifiers from the head are bifurcated earlier than those closer.}
Although
these are heuristics, this method has a relation to the linguistic
theory of center embedding of sentences.

\begin{figure}[t]
 \begin{minipage}[t]{0.30\linewidth}
 \centering \includegraphics[width = 1.\linewidth]{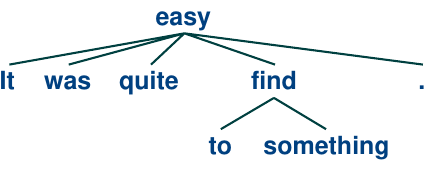}
 \subcaption{Dependency structure example}
 \end{minipage}
 \hspace{0.03\linewidth}
 \begin{minipage}[t]{0.30\linewidth}
 \centering \includegraphics[width = 1.\linewidth]{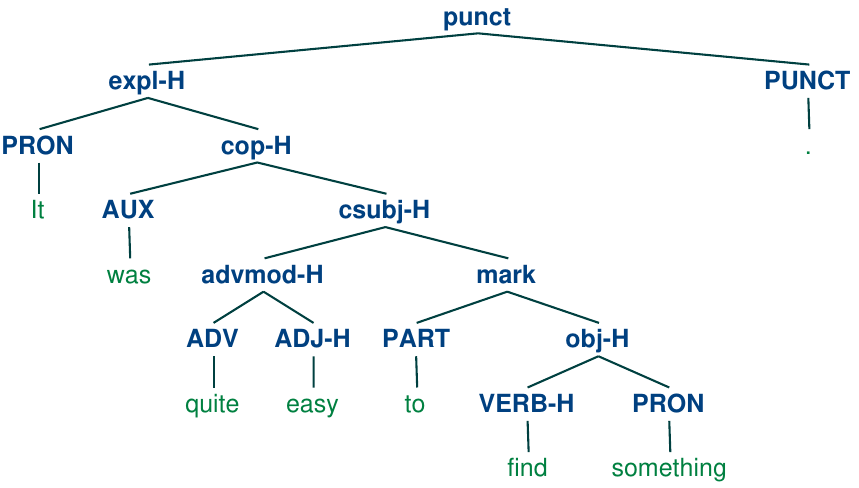}
 \subcaption{ Binary1 of (a), using grammar}
 \end{minipage}
 \hspace{0.03\linewidth}
 \begin{minipage}[t]{0.30\linewidth}
 \centering \includegraphics[width = 1.\linewidth]{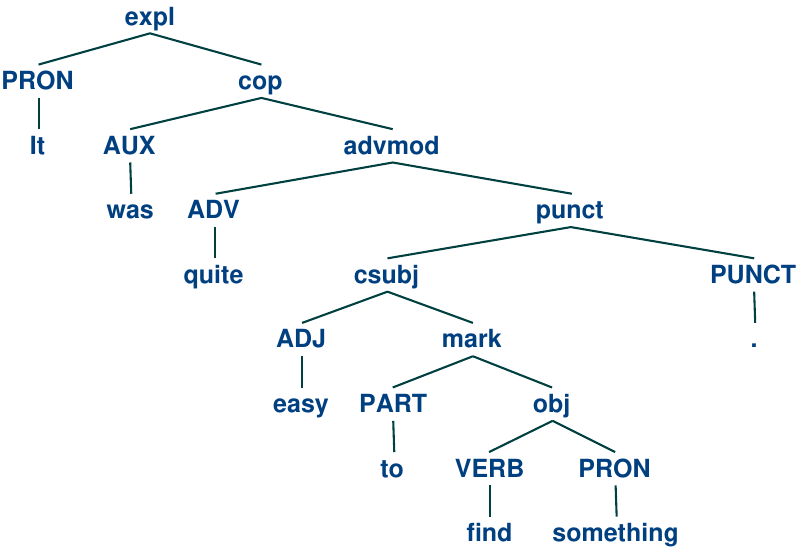}
 \subcaption{ Binary2 of (a), using heuristics}
 \end{minipage}
\caption{Dependency structure (a) and binary phrase structures (b, c)
  of "It was quite easy to find something." (a) is an example from the
  Universal Dependency Dataset \citep{ud_paper}. \label{samples}}
\end{figure}

A binarization example is shown in \figref{samples}, in which (a)
shows the tree of an original dependency structure, and (b) and (c)
show its binary-transformed phrase structure trees obtained with
Binary1 and 2, respectively.

As seen through these examples, the binarization methods each have
pros and cons. Binary1 has an advantage in that the resulting tree
structure reflects the correct sentence structure, but as mentioned
above, its applicability is limited. On the other hand, Binary2 does
not strictly reflect the sentence structure, but it is always
applicable. After application of Binary1 and 2, each tree's Strahler
number can be obtained by following the definition.

\subsection{Upper/Lower Limits of Strahler Number for Dependency Structures and Random Trees with $n$ Nodes: $R(n)$}
\label{sec:upperlower}
\label{sec:nrandom}
Binary1 and 2 are examples of possible methods for transforming a
dependency tree to a binary tree. Because the resulting Strahler
number depends on the resulting set of trees acquired via the
transformation method, we want to obtain the number's upper and lower
limits for all possible binary transformation methods.

In other words, a dependency tree $u_x$ can be transformed into
various binary trees by using some method under conditions that
reflect the original dependency structure. Let $U_x$ be the set of all
binarized trees for a given $u_x$, where each element is a binary tree
obtained with a particular binarization method. The upper and lower
limits are the maximum and minimum \blue{sizes} \red{of Strahler numbers}, respectively, in $U_x$.

The details of obtaining the upper/lower limits are described in
\appref{app:upperlower}, but we provide a summary here. A binarization
method constitutes a method to binarize each inner node $v$ of a tree
$t$. Binary1 and 2 are examples of different strategies, using a
grammar or heuristics. At each inner node, there is a binarization
method that maximizes or minimizes the Strahler number $S(v)$. The
maximizing method binarizes the subtree under $v$ so that it becomes
closer to a complete tree, whereas minimization makes the subtree
closer to a linear tree. We showed a very similar argument in
\secref{sec:maxminT2} for random trees. The maximum and minimum can be
calculated inductively to acquire the Strahler number's respective
upper and lower limits. Note that these limits are obtained while
ignoring the word order and the constraint of non-intersection,
because the maximum and minimum at each node $v$ are difficult to
compute under these constraints.

Thus far, we have explained how to acquire the upper/lower limits for
a particular tree $u_x$. We can also get the upper/lower limits across
the $u_x$ in a set of $U$. Specifically, for each subset $U(n)$ of
trees with $n$ nodes, the mean upper/lower limits of $U(n)$ can be
computed.

These upper and lower limits are comparable with those for the set of
random binary trees, $R_2(n)$, as mentioned in
\secref{sec:random}. Furthermore, apart from $R_2(n)$, we can consider
another set of random trees: all possible trees with $n$ nodes,
denoted as $R(n)$. The mean upper/lower limits of $R(n)$ for each $n$
are also empirically computable by the same method described in this
section. Because $|R(n)|$ is also a Catalan number
\citep{stanley2015catalan}, computation of the upper/lower limits of
$R(n)$ requires dynamic programming to cover the entire set. We
summarize that approach in Appendix B and give the details in
\appref{sec:average}.

\section{Data}
\label{sec:ud}
For the set $U$, as mentioned above, we use Universal Dependencies
\citep{ud,ud_paper}, version 2.8, to measure the
Strahler number for natural language. Universal Dependencies is a
well-known, large-scale project to construct large-scale annotated
data for natural language sentences. The annotation is defined under
the Universal Dependency scheme, which is a representation based on
dependency structure. The version used in this article contains 202
corpora across 114 languages. The corpora are listed in \blue{Table 1} \red{table 2} of
\appref{sec:datanumsentences}. Binary1 and 2 can be applied and upper
and lower limits can be calculated for all these data.

\section{Results}
\label{sec:exp}
To summarize the approach thus far, we have a dependency dataset $U$,
in which the subset of trees of size $n$ is denoted as $U(n)$. For
random trees, we have a set of binary random trees with $n$ leaf
nodes, denoted as $R_2(n)$, and a set of random trees with $n$ nodes,
$R(n)$.

As described above, for $R_2(n)$, the theoretical mean and upper and
lower bounds of the Strahler number are analytically known. For the
other sets, these values must be acquired empirically. For a tree $t$
belonging to one of those sets, we calculate the upper/lower limits of
Strahler numbers. In terms of $n$, the averages of each of these four
values can be acquired for $U(n)$ and $R(n)$. Binary1, 2 can also be
calculated for $U(n)$.

In this section, we consistently use color as follows. For $R_2(n)$,
we use black for the upper/lower limits and the mean, green for
Binary1, and blue for Binary2. For $U(n)$, we use pink for the upper
limit and red for the lower limit. For $R(n)$, we use purple for the
upper limit and orange for the lower limit.

\begin{table}[t]
\centering
\caption{Average $\pm$ standard deviation of the Strahler numbers for
  all dependency trees.}
\label{result_strahler_dist_table}
\begin{tabular}{cc}
\hline {} & All dependency trees \\ \hline Upper limits & $3.56 \pm
0.83$ \\ Lower limits & $2.71 \pm 0.60$ \\ Binary1 (with Grammar) &
$3.21 \pm 0.74$ \\ Binary2 (with heuristics) & $3.11 \pm 0.70$ \\
\end{tabular}
\end{table}

\begin{figure}[b]
\centering
 \begin{minipage}[t]{0.47\linewidth}
 \centering \includegraphics[width = 1.\linewidth]{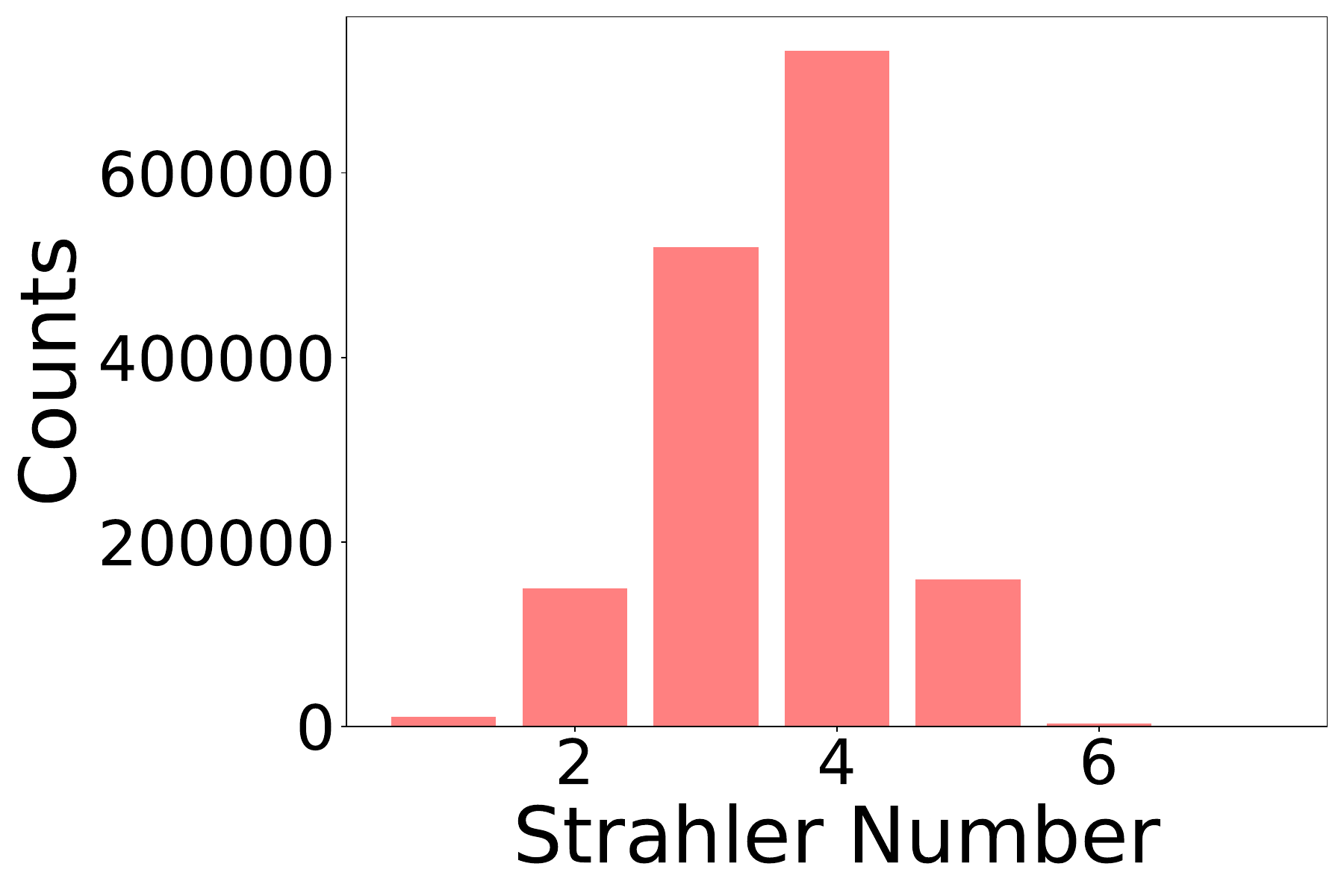}
 \subcaption{Upper limits}
 \end{minipage}
 \hspace{0.03\linewidth}
 \begin{minipage}[t]{0.47\linewidth}
 \centering \includegraphics[width = 1.\linewidth]{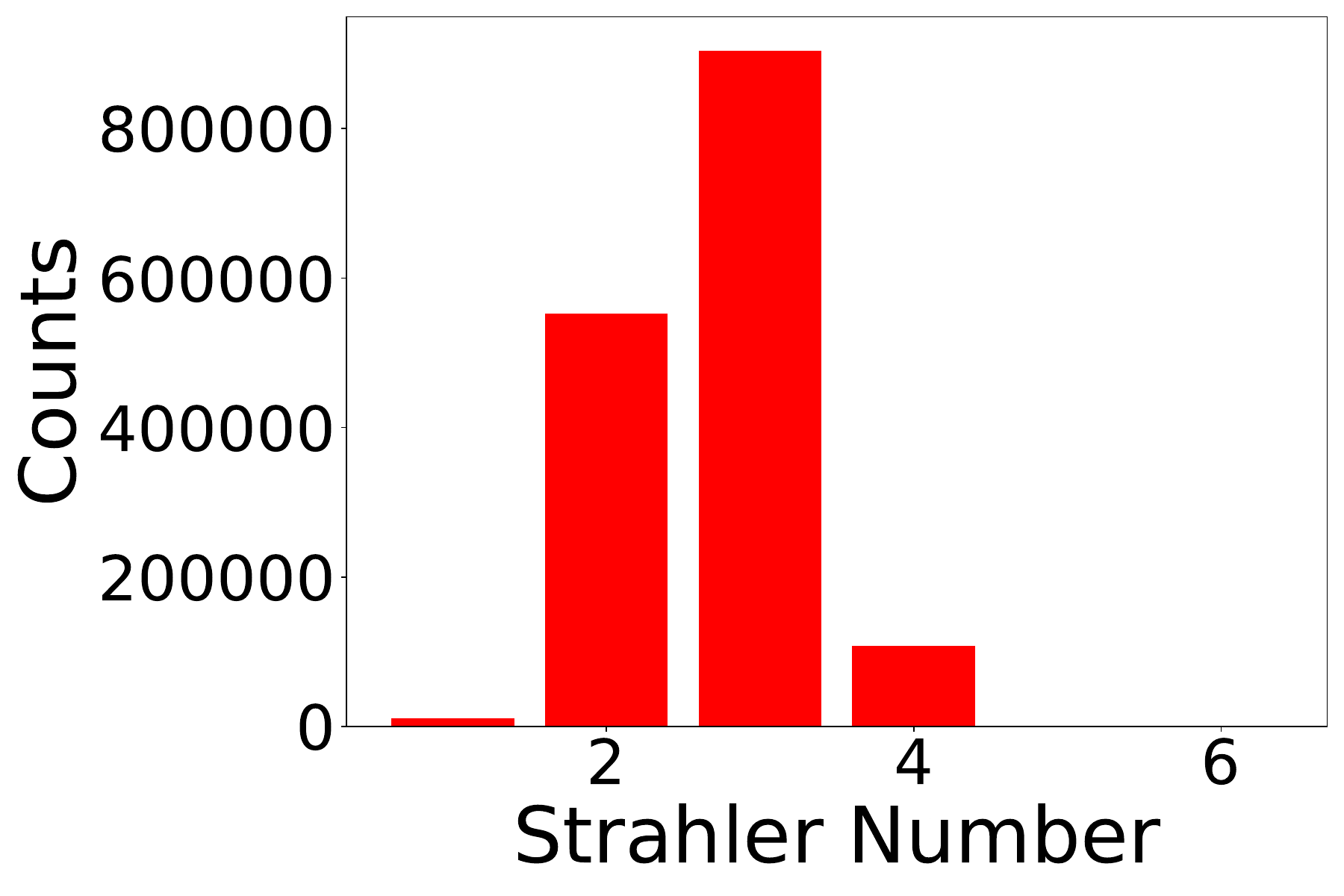}
 \subcaption{Lower limits}
 \end{minipage}\\
 \begin{minipage}[t]{0.47\linewidth}
 \centering \includegraphics[width =
   1.\linewidth]{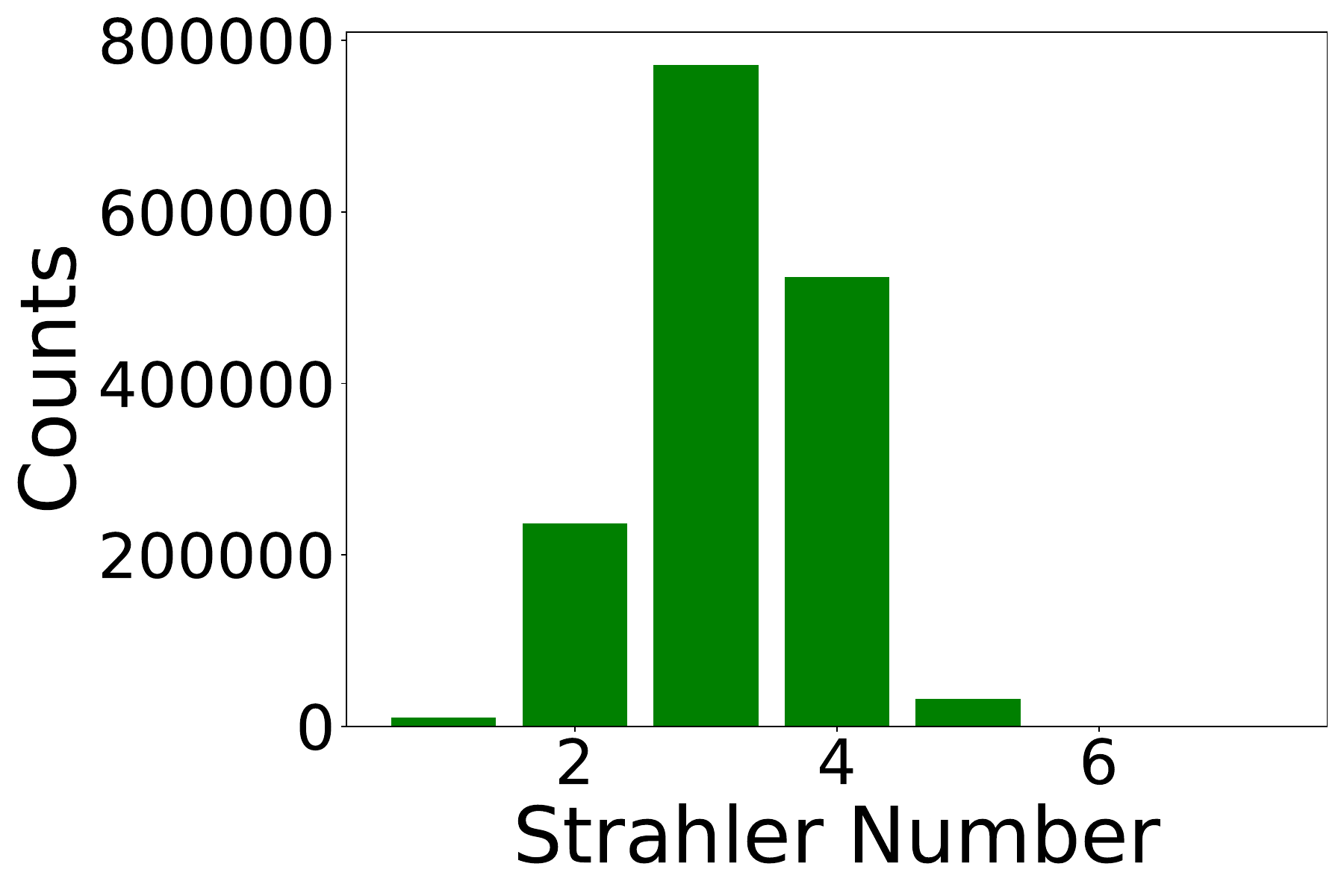} \subcaption{Binary1}
 \end{minipage}
 \hspace{0.03\linewidth}
 \begin{minipage}[t]{0.47\linewidth}
 \centering \includegraphics[width =
   1.\linewidth]{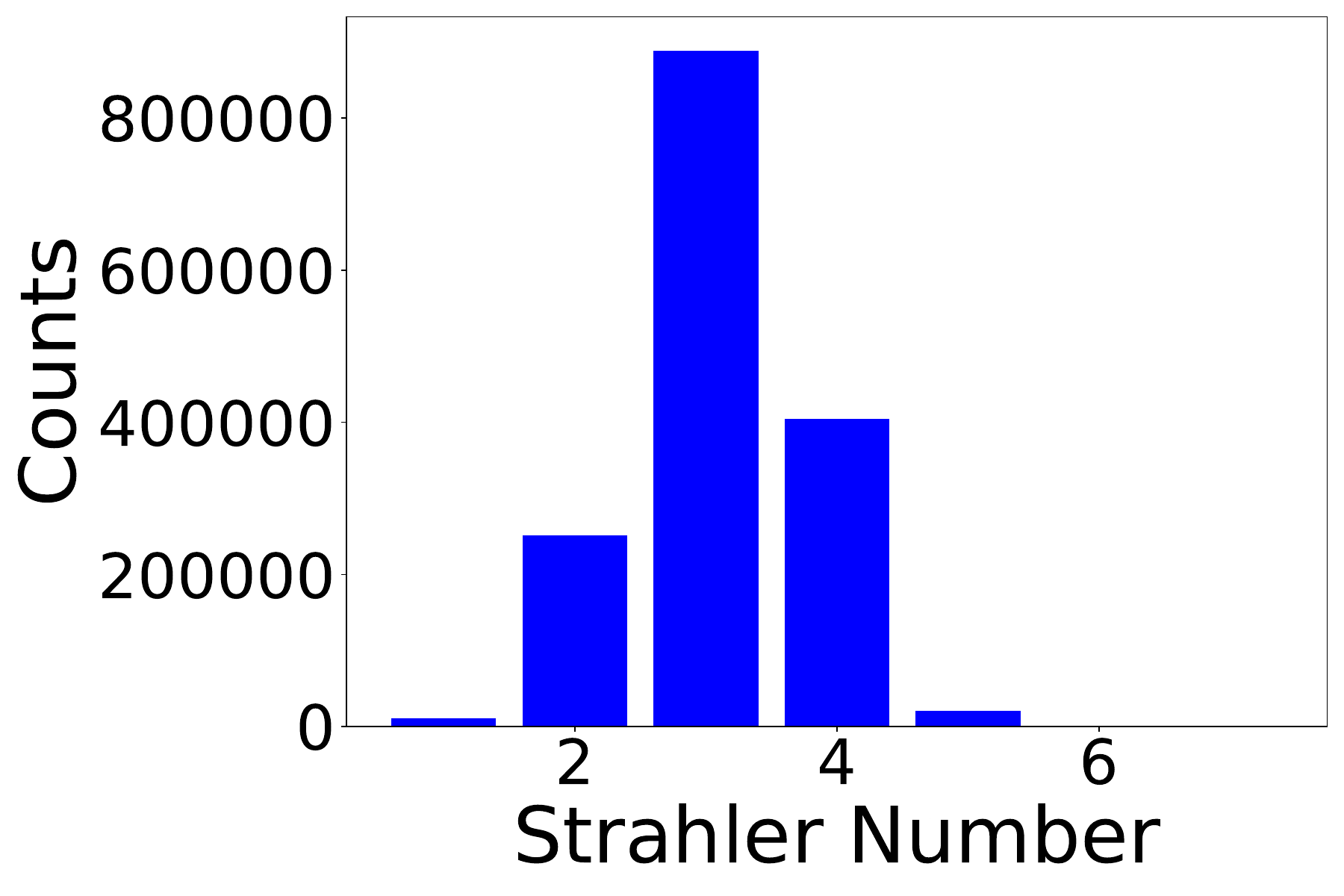} \subcaption{Binary2}
 \end{minipage}
\caption{Histogram of the Strahler numbers of dependency trees.}
 \label{result_strahler_dist}
\end{figure}

\subsection{Strahler Number of Sentence Structure}
\label{sec:strahler_dist}
\tabref{result_strahler_dist_table} lists the means and standard
deviations for the entire dependency dataset. We see that the
average Strahler
number of a dependency structure is usually less than 4. The Binary1
and Binary2 values are between the upper and lower limits. For each
corpus, the specific means and standard deviations for Binary1 and 2
and the upper/lower limits for $U(n)$ are listed in Appendices
E.2-E.5, Tables 3-10.

\figref{result_strahler_dist} shows a histogram of the Strahler
numbers. It can be seen that the distribution shifts from large to
small in the order of the upper limit, Binary1, Binary2, and the lower
limit. Note that Binary1 and 2 show pretty similar results, regardless
of the binarization method. The median Strahler number is 4 for the
upper limit, and 3 for all other cases. Strahler numbers larger than 4
are clearly very scarce.

The dependency dataset includes data of various language groups,
genres, and modes (speech/writing). According to our analysis, the
differences with respect to Strahler number among datasets are not distinct across this variety of
data. The largest Strahler number is 7, and the smallest is
1. Examples of both extremes are given in \appref{sec:extremes}. The
examples with a number of 1 are mainly one-word salutations,
interjections, and names (even without periods; \appref{sec:extremes},
Table 11). On the other hand, sentence examples with a Strahler number
of 7 are very rare and contain a large number of words. As seen here,
sentences with a larger Strahler number above 4 are atypical and
include examples for which it might be questionable to call them
sentences. The dependency dataset includes such questionable entries,
and the Strahler number could provide evidence to quantify such
irregularities in the corpora.

\subsection{Growth of Strahler Number w.r.t. Sentence Length}
\label{sec:strahler_sentence_length}
Originally, when the Strahler number was used for analysis of rivers
in England, it was found to be 4. We can also conclude from the
previous section that the Strahler number for sentence structure is
almost 4. This leads us to wonder how this number is significant. Thus
far, we have discussed the Strahler number as a constant value with a
given distribution. In the following, we show that it is not a
constant but merely {\em looks} like a constant, because it grows
slowly with respect to $n$, and the range of sentence lengths is
limited. In fact, the number depends on the logarithm of the tree size
$n$.

\begin{figure}[t]
\centering \includegraphics[width =1.\linewidth]{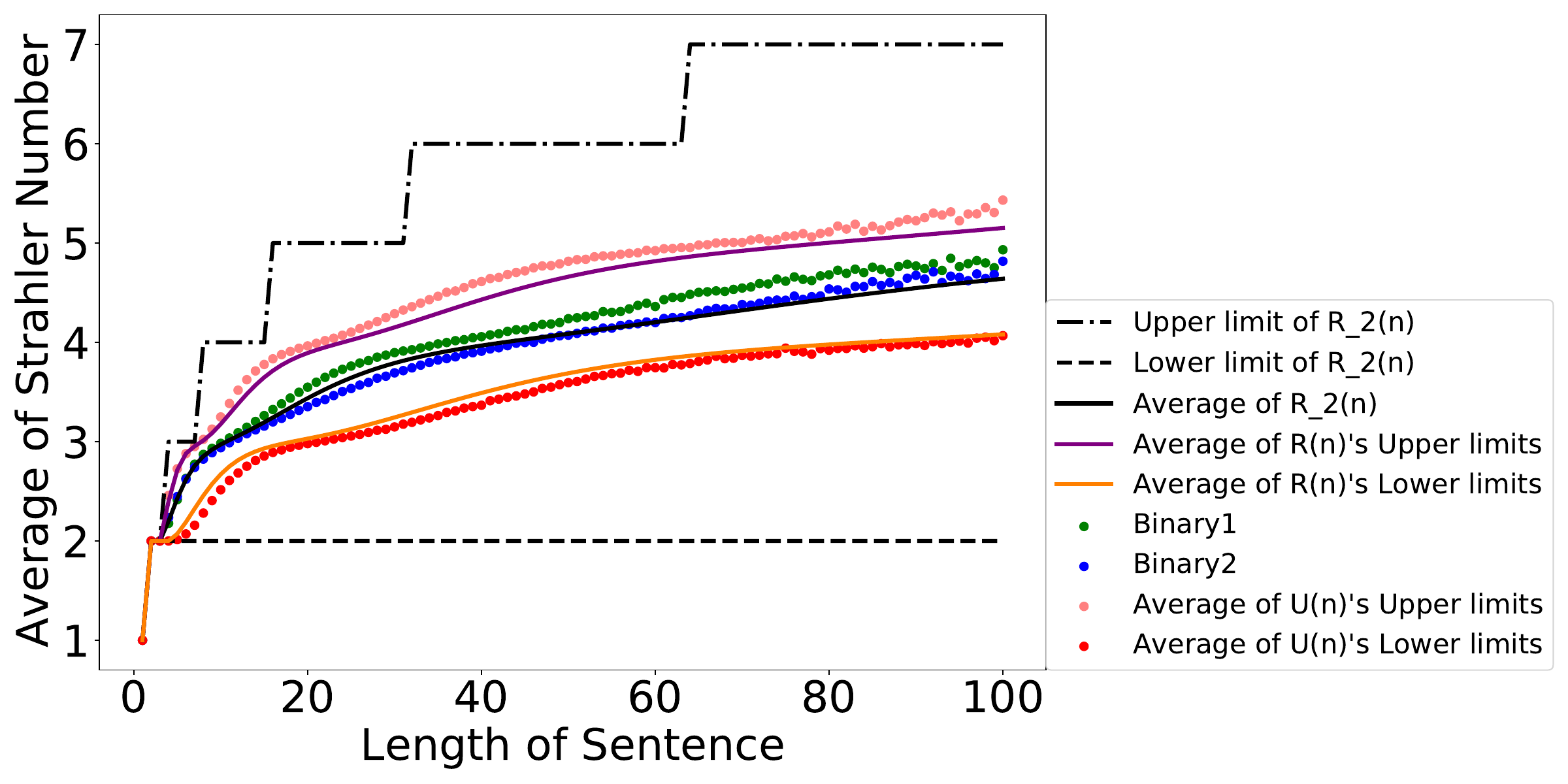}
\caption{Average Strahler number with respect to the tree size $n$.}
\label{result_strahler_length}
\end{figure}

\figref{result_strahler_length} shows the mean results for the tree
sets $U(n)$, $R_2(n)$, and $R(n)$, as summarized at the beginning of
this section. The black analytical lines for $R_2(n)$ indicate the
exact values following the theory explained in
\secref{sec:random}. For the other sets, the plots show empirical
results measured across trees of size $n$. All plots approximately
increase logarithmically, but none of them are smooth, as they have a
step at $n=2$, and they globally fluctuate by changing their
logarithmic base. Overall, the necessary number of stack spaces is
bounded by the logarithm of the tree size.

For each $n$, the possible range of Strahler numbers for $R_2(n)$,
which extends between the upper and lower black lines, is obviously
far wider than the range for $U(n)$. On the other hand, the range for
$U(n)$ is between the pink and red points. The range for $R(n)$ is
between the purple and orange lines, which is slightly narrower than
the range for $U(n)$, despite $R(n)$ being the average of all {\em
  random} trees with $n$ nodes.

\begin{figure}[t]
\centering \includegraphics[width = 1.\linewidth]{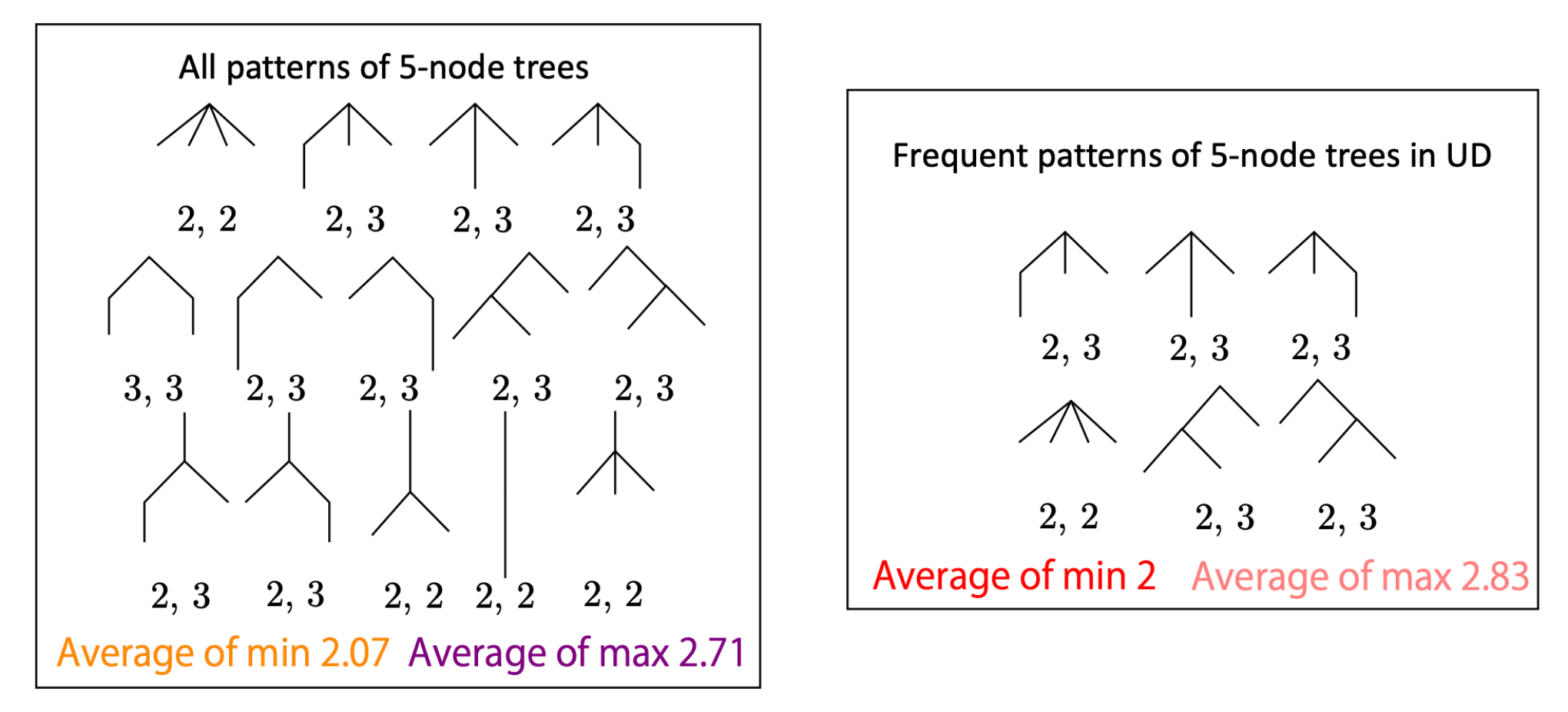}
\caption{Sample trees for $n=5$: (left) all possible trees, and
  (right) tree structures appearing frequently among Universal Dependency
  (UD) trees. Two numbers beneath each tree indicate the mininum and
  maximum, respectively, of the Strahler number across all possible binarialization methods.}
\label{fig:n5}
\end{figure}

These results can be understood from a small example. \figref{fig:n5}
shows a set of trees of size $n=5$, with all such possible trees on
the left, and typical structures appearing frequently in the
dependency dataset on the right. The distribution of tree shapes in
the dataset varies, with the set of trees on the right accounting for
80\% of the total. The upper and lower limits of the Strahler number
are listed below each tree. The averages are listed at the bottom of
each box in the corresponding colors from the scheme used throughout
this section. For $R(n)$, the respective upper/lower limits are 2.71
and 2.07; in contrast, if the six trees on the right appeared equally,
the upper/lower limits would be 2.83 and 2. Thus, the range of $R(n)$
is narrower than that of $U(n)$, even in this small sample with $n=5$.

The actual plots in \figref{result_strahler_length} were obtained by
computing the average across the distribution of shapes, but the range
of $R(n)$ is still contained within that of $U(n)$. This small example
with $n=5$ explains why the range of $R(n)$ can be almost the same or
even smaller than that of $U(n)$: the Strahler number is mostly the
same for any kind of tree of the same size and does not especially
characterize the tree shape.

\begin{figure}[t]
 \centering \includegraphics[width=0.47\linewidth]{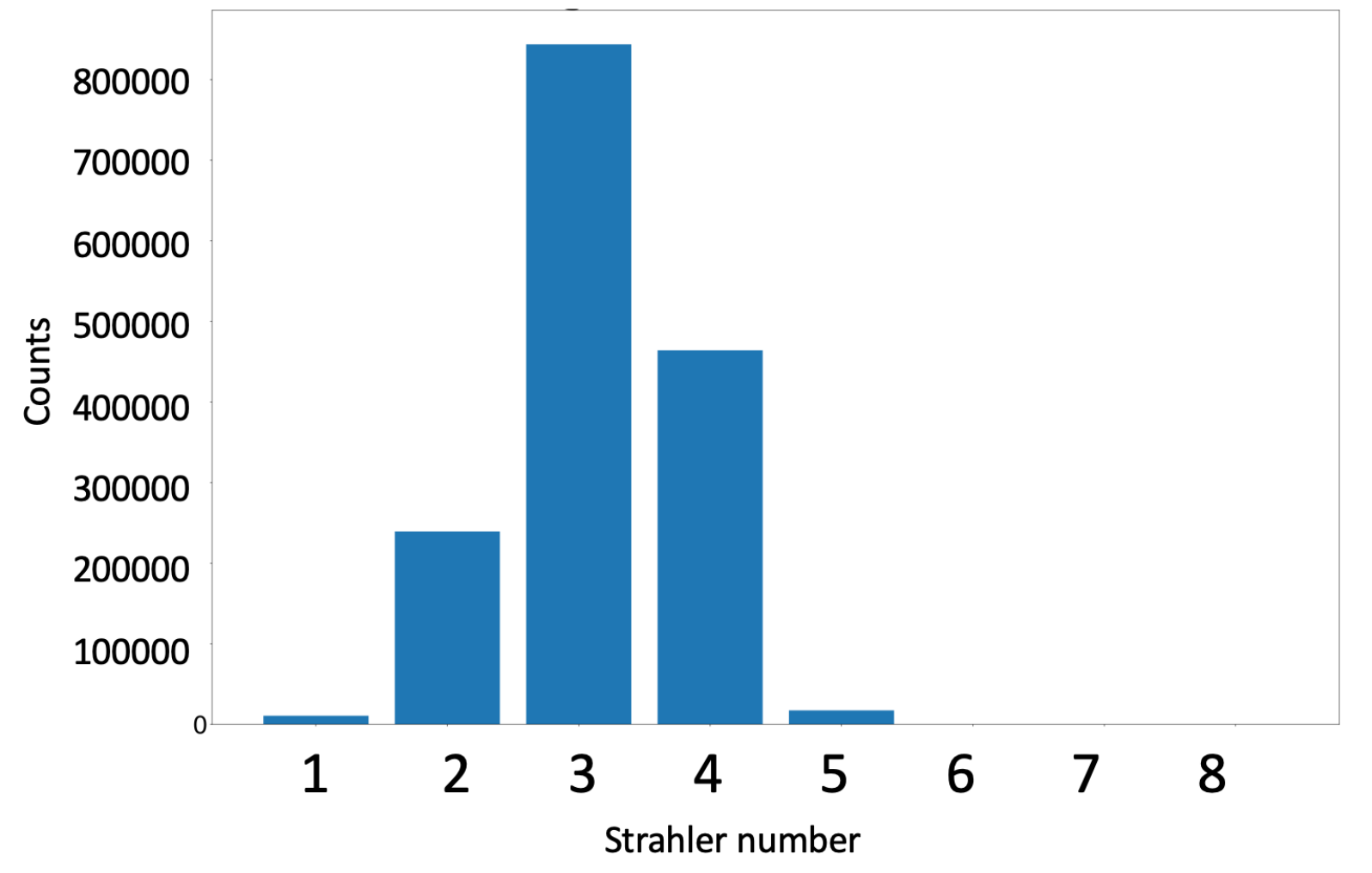}
 \caption{Histogram of the Strahler numbers of random trees from
   $R_2(n)$, when sampling $n$ from the distribution of Binary
   1. \label{fig:random}}
\end{figure}

To show how the Strahler numbers of natural language trees are not
much different from those of random trees, we can examine a histogram
of the results for $R_2(n)$, the null model. Because any arbitrary
large tree of size $n$ can be generated for $R_2(n)$ and the Strahler
number grows with respect to $n$, trees lengths were sampled from
Binary1, and for each $n$, a Strahler number was sampled from the
population of $R_2(n)$. \figref{fig:random} shows a histogram of
$R_2(n)$ for the total number of sentences in UD. In
comparison with Figure 6(c), there are more trees here with a Strahler
number of 3. The Strahler number's average and standard deviation were
$3.15 \pm 0.70$ here, a little smaller than for Binary1. Thus, the
results do not show much difference between random trees and natural
language trees.

\begin{figure}[t]
\centering \includegraphics[width
  =\linewidth]{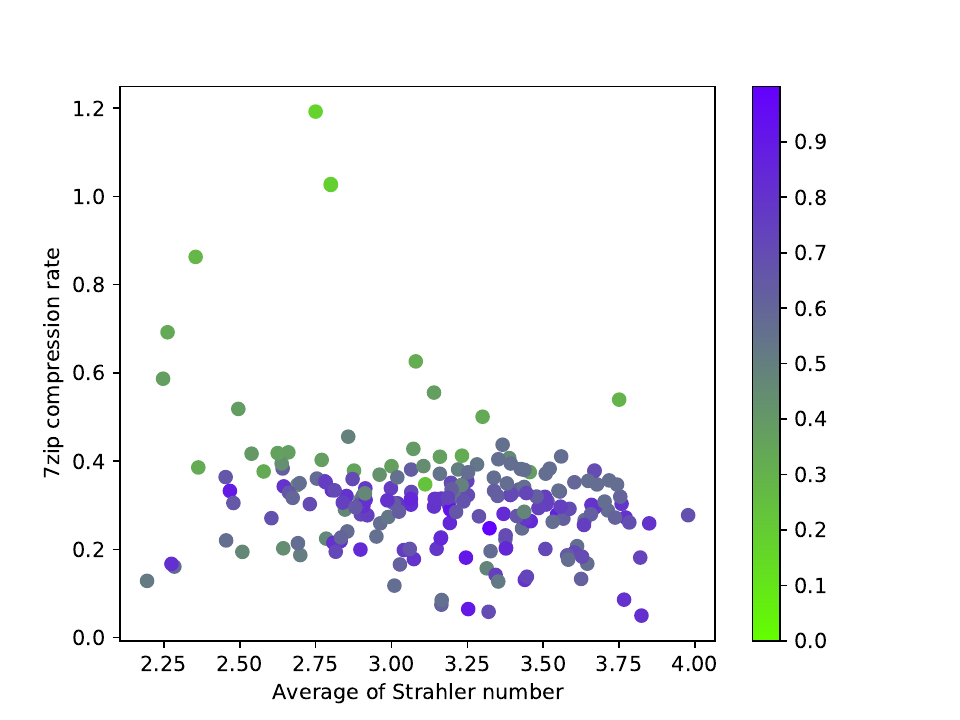}
\caption{7zip compression rate with respect to the average Strahler
  number per corpus for Binary1.}
\label{fig:7zip}
\end{figure}

\subsection{Difference of Strahler Number from Compression Factor}

To investigate the Strahler number's significance further, this section
demonstrates its originality in terms of how it differs from other
complexity measures.

First, although the Strahler number depends on the logarithm of the
tree size $n$, the depth and Strahler number are
different. Previously, a tree's depth was considered as a kind of tree
complexity, as in \citet{yngve60}, but a tree's depth does not
necessarily signify sentence complexity. For example, a linear tree of
size $n$ has a deptch of $n$, yet its Strahler number
is \blue{1} \red{2}. The Strahler number is a much more sophisticated
statistic related to the amount of necessary memory, as argued above
in this article. Theoretically, the Strahler number's dependence
almost on the logarithm of the tree size is not trivial, as
demonstrated in \citet{flajolet79}'s analytical work. Experimentally,
the Strahler number requires actual, nontrivial calculation, which has
been the main point of this work.

Another representative complexity measure that has frequently been
applied to quantify natural language text is the Shannon entropy
\citep{shannon48}. Other, entirely different characteristics could
have some correlation with the Shannon entropy. For example,
\cite{cl19} showed how Taylor's exponent, which quantifies fluctuation, has
an empirical relation with the perplexity, another form of the Shannon
entropy. The remainder of this section shows that the Strahler number
does not correlate with the Shannon entropy and thus captures a
different aspect of natural language complexity.

\cite{cl15} showed how direct calculation of the Shannon entropy
depends on the corpus length if it is calculated via the plug-in
entropy, $-\sum_i p(w_i) \log (w_i)$, with the $w_i$ being elements,
i.e., either words or characters. This dependence on the corpus length
is alleviated by using a good compression method $\gamma$ to calculate
the compression rate $r = |D|/|\gamma(D)|$, where $D$ is the original
corpus, $\gamma(D)$ is the compressed version, and $|\cdot|$ denotes a
sequence's length. The compression rate $r$ is theoretically proven to
tend to the true Shannon entropy, provided that $\gamma$ is universal,
and that $D$ is stationary and ergodic
\citep{coverthomas}. \cite{friedrich21} applied the same strategy to
compress texts and thus examine the complexity of domain-specific
texts. Furthermore, \cite{entropy16} showed that, among different
compression methods, the PPM \red{(Prediction by Partial Match)} method \citep{ppm} behaves correctly
\red{following theory} at
least for some random data with a much faster compression speed
relative to the text length than other popular methods such as
Lempel-Ziv.

Given previous reports above, for every corpus in Binary1, 
we compared the average of the Strahler number and the PPM
compression rate\footnote{We used the p7zip@16.02\_5 package.} for
text sequences taken only from grammatically annotated
corpora.\footnote{We excluded the following corpora, which provide
sentence structures without words: UD\_Hindi\_English-HIENCS,
UD\_Arabic-NYUAD, UD\_Japanese-BCCWJ, UD\_English-ESL, UD\_French-FTB,
UD\_English-GUMReddit, UD\_Mbya\_Guarani-Dooley.}

\figref{fig:7zip} shows the results. Every point corresponds to a UD
corpus, with the horizontal axis indicating the average Strahler
number, and the vertical axis indicating the compression rate. The
values on the horizontal axis correspond to the values mentioned in
Appendix E.2. To illustrate the influence of \red{the corpus length on
  the compression rate}, the points have colors ranging from blue (or
purple) to green in terms of the ratio of the logarithm of each
corpus' number of sentences to the logarithm of the number of
sentences in the largest corpus of UD. As seen in the figure, the
green points (representing very small corpora) are located toward the
top, whereas the majority of the blueish (purplish) points (larger
corpora) are not necessarily correlated with their vertical
locations. Thus, we believe that the text length's influence on the
compression rate is insignificant. Note that if we directly use the
Shannon plug-in entropy, the length does influence the location,
because the UD corpora are limited in size.

As seen in \figref{fig:7zip}, the compression rate is scattered almost the same
for any Strahler number. This shows that the Strahler number
\red{is independent of the compression rate, and} 
captures
a different aspect of the complexity of language structure than the
Shannon entropy\red{.} \blue{: they are independent}. Whereas the Shannon entropy
highlights the complexity over a sequence, the Strahler number
indicates the complexity of the memory required to process a tree 
structure, and this figure shows how these two measures are
essentially different. We consider this article's value to lie in
investigating the Strahler number's behavior as known in statistical
physics while revealing another aspect of the complexity of natural
language.

\section{Discussion}
Previous reports on the maximum number of short-term memory areas that
can be cognitively used have consistently suggested a value of 3 to
5. In a survey of previous works, \cite{cowan01} summarized
cognitive works that tested the maximum number of events or instances
that could be remembered through human psychological experiments,
e.g., via instant memory \citep{Sperling1960} or graphics
\citep{LuckandVogel1997}. Cowan stated that the number of such memory
areas is 3 to 5 and referred to the value as a ``magical number.''

The possible relation of such a maximum number of short-term or local
memory areas to the number of memory areas required for sentence
understanding is nontrivial. Memory is necessary to understand
sentences, and one model for theorizing this could be based on the
shift-reduce approach, as presented in this work. Under this setting, the experimental results in
this article shows that this number is within the range of Cowan's
{\em magical number}.

In parallel to these findings in psychology, there have been works on
parsing sentences, as mentioned in the Related Work section. Some
works reported the maximum number of memory areas to process a
sentence. Although the number depends on the parsing method, the works
since \blue{\cite{abney1991memory},} \cite{schuler2010broad} have reported the number
to be three to four, as we have found here.

Hence, our work's contribution is reasoning about this magical number
in a rigorous setting via the lower limits of shift-reduce evaluation
of a tree. Although our findings are limited to this setting, we have
shown that the Strahler number grows with the logarithm of the
sentence length. Conjecturing on the similarity and distinction of
parsing techniques and our work, we predict that the lower limit of
the memory amount necessary to process human sentences by parsing also
grows logarithmically. Then, the reason why this number is three to
four lies in the range of sentence lengths: because a logarithmic
function grows slowly, a large range falls into a logarithmic value of
three to four.

Furthermore, through a comparison with random trees from $R(n)$, we
have shown that this result is not specific to human sentences but
applies in general to a wider set of all possible random tree
shapes. This understanding suggests the possibility that the magical
number is not specific to human-generated trees but could derive from
the behavior of trees in general.

\section{Conclusion}
In this article, we examined the use of the Strahler number to
understand the nature of sentence structure. The Strahler number was
originally defined to analyze the complexity of river
bifurcation. Here, we applied it to sentences, which is the first such
application to the best of our knowledge. Because the tree structure
dataset used here is much larger than the datasets used in previous
applications, it enabled a statistical study of the Strahler number in
comparison with random trees.

The Strahler number entails the memory necessary to process a sentence
structure via the basic shift-reduce method. We proposed ways to
compute a sentence's Strahler number, via two binarization methods and
the upper and lower limits across all possible binarization
methods. The experimental results showed that the Strahler number of a
natural language sentence structure is almost 3 or 4. This number was
found to grow with the sentence length, and the upper/lower limits
were found to be close to those of random trees of the same length,
which is the Strahler number's key characteristic with respect to
trees, including random trees. These findings provide evidence and
understanding of the memory limit discussed to date in relation to the
``magical number'' in psychology and the upper limit on the memory
necessary to parse texts.

\section*{Acknowledgement}

This work was supported by JST, CREST Grant Number JPMJCR2114, Japan. 

\bibliographystyle{natbib} \bibliography{strahler}

\begin{thebibliography}{}

\bibitem[Abney and Johnson(1991)]{abney1991memory}
Abney, S.~P. and Johnson, M. (1991).
\newblock {Memory requirements and local ambiguities of parsing strategies}.
\newblock {\em Journal of Psycholinguistic Research}, {\bf 20}, 233--250.

\bibitem[Altmann {\em et~al.}(2009)]{altmann2009}
Altmann, E.~G., Pierrehumbert, J.~B., and Motter, A.~E. (2009).
\newblock Beyond word frequency: Bursts, lulls, and scaling in the temporal
  distributions of words.
\newblock {\em PLoS One}, {\bf 4}(e7678).

\bibitem[Altmann {\em et~al.}(2012)]{Altmann2012}
Altmann, E.~G., Cristadoro, G., and Esposti, M.~D. (2012).
\newblock On the origin of long-range correlations in texts.
\newblock {\em Proceedings of the National Academy of Sciences}, {\bf 109}(29),
  11582--11587.

\bibitem[Auber {\em et~al.}(2004)]{auber04}
Auber, D., Domenger, J.-P., Delest, M., Duchon, P., and F{\'e}dou, J.-M.
  (2004).
\newblock {New Strahler Numbers for Rooted Plane Trees}.
\newblock In M.~Drmota, P.~Flajolet, D.~Gardy, and B.~Gittenberger, editors,
  {\em Mathematics and Computer Science III}, pages 203--215, Basel.
  Birkh{\"a}user Basel.

\bibitem[Beer and Borgas(1993)]{beer93}
Beer, T. and Borgas, M. (1993).
\newblock {Horton's Laws and the fractal nature of streams}.
\newblock {\em Water Resources Research}, {\bf 29}(5), 1475--1487.

\bibitem[Bell {\em et~al.}(1990)]{ppm}
Bell, T., Cleary, J., and Witten, I. (1990).
\newblock {\em Text Compression}.
\newblock Prentice Hall.

\bibitem[Buchholz(2002)]{buchholz02}
Buchholz, S.~N. (2002).
\newblock {\em {Memory-Based Grammatical Relation Finding}}.
\newblock Eigen beheer.
\newblock Doctoral Thesis.

\bibitem[Buk and Rovenchak(2007)]{buk07}
Buk, S. and Rovenchak, A. (2007).
\newblock Menzerah-altmann law for syntactic structures in ukrainian.
\newblock {\tt https://arxiv.org/pdf/cs/} {0701194.pdf}.

\bibitem[Chomsky(1956)]{Chomsky56}
Chomsky, N. (1956).
\newblock {Three models for the description of language}.
\newblock {\em IRE Transactions on Information Theory}, {\bf 2}(3), 113--124.

\bibitem[Cover and Thomas(1991)]{coverthomas}
Cover, T.~M. and Thomas, J.~A. (1991).
\newblock {\em Elements of Information Theory}.
\newblock John Wiley \& Sons, Inc.

\bibitem[Cowan(2001)]{cowan01}
Cowan, N. (2001).
\newblock {The magical number 4 in short-term memory: A reconsideration of
  mental storage capacity}.
\newblock {\em Behavioral and Brain Sciences}, {\bf 24}, 87 -- 114.

\bibitem[de~Marneffe {\em et~al.}(2021)]{ud}
de~Marneffe, M.-C., Manning, C.~D., Nivre, J., and Zeman, D. (2021).
\newblock {Universal Dependencies}.
\newblock {\em Computational Linguistics}, pages 1--52.

\bibitem[Degiuli(2019a)]{degiuli19-2}
Degiuli, E. (2019a).
\newblock Emergence of order in random languages.
\newblock {\bf 52}.
\newblock 504001.

\bibitem[Degiuli(2019b)]{degiuli19-1}
Degiuli, E. (2019b).
\newblock Random language model.
\newblock {\bf 122}.
\newblock 128301.

\bibitem[Ebeling and Neiman(1995)]{ebeling1995}
Ebeling, W. and Neiman, A. (1995).
\newblock Long-range correlations between letters and sentences in texts.
\newblock {\em Physica A}, {\bf 215}, 233--241.

\bibitem[Ebeling and {P\"{o}schel}(1993)]{ebeling1994}
Ebeling, W. and {P\"{o}schel}, T. (1993).
\newblock Entropy and long-range correlations in literary {E}nglish.
\newblock {\em Europhysics Letters}, {\bf 26}(4), 241--246.

\bibitem[Ehrenfeucht {\em et~al.}(1981)]{ehrenfeucht1981etol}
Ehrenfeucht, A., Rozenberg, G., and Vermeir, D. (1981).
\newblock {On Etol Systems with Finite Tree-Rank}.
\newblock {\em SIAM Journal on Computing}, {\bf 10}(1), 40--58.

\bibitem[Ershov(1958)]{Ershov58}
Ershov, A.~P. (1958).
\newblock {On Programming of Arithmetic Operations}.
\newblock {\em Commun. ACM}, {\bf 1}(8), 3--6.

\bibitem[Fern{\'a}ndez-Gonz{\'a}lez and
  G{\'o}mez-Rodr{\'\i}guez(2019)]{fernandez2019faster}
Fern{\'a}ndez-Gonz{\'a}lez, D. and G{\'o}mez-Rodr{\'\i}guez, C. (2019).
\newblock {Faster shift-reduce constituent parsing with a non-binary, bottom-up
  strategy}.
\newblock {\em Artificial Intelligence}, {\bf 275}, 559--574.

\bibitem[Fern{\'a}ndez-Gonz{\'a}lez and
  G{\'o}mez-Rodr{\'\i}guez(2023)]{fernandez2023discontinuous}
Fern{\'a}ndez-Gonz{\'a}lez, D. and G{\'o}mez-Rodr{\'\i}guez, C. (2023).
\newblock {Discontinuous grammar as a foreign language}.
\newblock {\em Neurocomputing}, {\bf 524}, 43--58.

\bibitem[Fern\'{a}ndez-Gonz\'{a}lez and Martins(2015)]{p15-1147}
Fern\'{a}ndez-Gonz\'{a}lez, D. and Martins, A. F.~T. (2015).
\newblock {Parsing as Reduction}.
\newblock In {\em Proceedings of the 53rd Annual Meeting of the Association for
  Computational Linguistics and the 7th International Joint Conference on
  Natural Language Processing Volume 1 : Long Papers}, pages 1523--1533.
  Association for Computational Linguistics.

\bibitem[Fischer {\em et~al.}(2021)]{fischer21}
Fischer, M., Herbst, L., Kuhn, L., and Wicke, K. (2021).
\newblock {Tree balance indices: a comprehensive survey}.
\newblock https://arxiv/abs/2109.12281.

\bibitem[Flajolet {\em et~al.}(1979)]{flajolet79}
Flajolet, P., Raoult, J., and Vuillemin, J. (1979).
\newblock {The number of registers required for evaluating arithmetic
  expressions}.
\newblock {\em Theoretical Computer Science}, {\bf 9}(1), 99--125.

\bibitem[Forster(1968)]{forster1968sentence}
Forster, K.~I. (1968).
\newblock {Sentence completion in left-and right-branching languages}.
\newblock {\em Journal of Verbal Learning and Verbal Behavior}, {\bf 7}(2),
  296--299.

\bibitem[Friedrich(2021)]{friedrich21}
Friedrich, R. (2021).
\newblock Complexity and entropy in legal language.
\newblock {\em Frontiers in Physics}, {\bf 9}.
\newblock 671882.

\bibitem[Gibson(2000)]{gibson00}
Gibson, E. (2000).
\newblock {The dependency locality theory: A distance-based theory of
  linguistic complexity}.
\newblock {\em Image, Language, Brain: Papers from the First Mind Articulation
  Project Symposium}, pages 94--126.

\bibitem[Grenander {\em et~al.}(2022)]{grenander2022sentence}
Grenander, M., Cohen, S.~B., and Steedman, M. (2022).
\newblock {Sentence-Incremental Neural Coreference Resolution}.
\newblock In {\em Proceedings of the 2022 Conference on Empirical Methods in
  Natural Language Processing}, pages 427--443.

\bibitem[{Horton}(1945)]{horten45}
{Horton}, R.~E. (1945).
\newblock {Erosional Development of Streams and Their Drainage Basins;
  Hydrophysical Approach to Quantitative Morphology}.
\newblock {\em Geological Society of America Bulletin}, {\bf 56}(3), 275.

\bibitem[Hou {\em et~al.}(2017)]{hou17}
Hou, R., Huang, C.-R., San~Do, H., and Liu, H. (2017).
\newblock A study on correlation between chinese sentence and constituting
  clauses based on the menzerath-altmann law.
\newblock {\em Journal of Quantitative Linguistics}, pages 350--366.

\bibitem[Kimball(1975)]{Kimball75}
Kimball, J.~P. (1975).
\newblock {\em {Predictive Analysis and Over-the-Top Parsing}}.
\newblock Brill, Leiden, The Netherlands.

\bibitem[Kobayashi and Tanaka-Ishii(2018)]{acl18}
Kobayashi, T. and Tanaka-Ishii, K. (2018).
\newblock {Taylor}'s law for human linguistic sequences.
\newblock {\em Annual Conference of Association for Computational Lingusitics},
  pages 1138--1148.

\bibitem[Kong {\em et~al.}(2015)]{n15-1080}
Kong, L., Rush, A.~M., and Smith, N.~A. (2015).
\newblock {Transforming Dependencies into Phrase Structures}.
\newblock In {\em Proceedings of the 2015 Annual Conference of the North
  American Chapter of the Association for Computational Linguistics:Human
  Language Technologies}, pages 788--798. Association for Computational
  Linguistics.

\bibitem[Lin and Tegmark(2017)]{lin_2016}
Lin, H.~W. and Tegmark, M. (2017).
\newblock Critial behavior in physics and probabilistic formal languages.
\newblock {\em Entropy}, {\bf 19}(7), 299.

\bibitem[Liu(2008)]{liu2008dependency}
Liu, H. (2008).
\newblock {Dependency Distance as a Metric of Language Comprehension
  Difficulty}.
\newblock {\em Journal of Cognitive Science}, {\bf 9}(2), 159--191.

\bibitem[Liu {\em et~al.}(2017)]{liu2017dependency}
Liu, H., Xu, C., and Liang, J. (2017).
\newblock {Dependency distance: A new perspective on syntactic patterns in
  natural languages}.
\newblock {\em Physics of life reviews}, {\bf 21}, 171--193.

\bibitem[Luck and Vogel(1997)]{LuckandVogel1997}
Luck, S.~J. and Vogel, E.~K. (1997).
\newblock {The capacity of visual working memory for features and
  conjunctions}.
\newblock {\em Nature}, {\bf 390}(6657), 279--281.

\bibitem[Ma{\v{c}}utek {\em et~al.}(2017)]{macutek17}
Ma{\v{c}}utek, J., {\v{C}}ech, R., and Mili{\v{c}}ka, J. (2017).
\newblock Menzerath-altmann law in syntactic dependency structure.
\newblock In {\em Proceedings of the Fourth International Conference on
  Dependency Linguistics (Depling 2017)}, pages 100--107, Pisa,Italy.
  Link{\"o}ping University Electronic Press.

\bibitem[Miller(1956)]{miller56}
Miller, G.~A. (1956).
\newblock {The magical number seven, plus or minus two: Some limits on our
  capacity for processing information.}
\newblock {\em Psychological review}, {\bf 63}(2), 81.

\bibitem[Nivre {\em et~al.}(2020a)]{Nivre20ud}
Nivre, J., de~Marneffe, M.-C., Ginter, F., Haji{\v{c}}, J., Manning, C.~D.,
  Pyysalo, S., Schuster, S., Tyers, F., and Zeman, D. (2020a).
\newblock {{U}niversal {D}ependencies v2: An Evergrowing Multilingual Treebank
  Collection}.
\newblock In {\em Proceedings of the 12th Language Resources and Evaluation
  Conference}, pages 4034--4043, Marseille, France. European Language Resources
  Association.

\bibitem[Nivre {\em et~al.}(2020b)]{nivre20universal}
Nivre, J., de~Marneffe, M.-C., Ginter, F., Haji{\v{c}}, J., Manning, C.~D.,
  Pyysalo, S., Schuster, S., Tyers, F., and Zeman, D. (2020b).
\newblock {{U}niversal {D}ependencies v2: An Evergrowing Multilingual Treebank
  Collection}.
\newblock In {\em Proceedings of the Twelfth Language Resources and Evaluation
  Conference}, pages 4034--4043, Marseille, France. European Language Resources
  Association.

\bibitem[Nivre {\em et~al.}(2020c)]{ud_paper}
Nivre, J., de~Marneffe, M.-C., Ginter, F., Haji{\v{c}}, J., Manning, C.~D.,
  Pyysalo, S., Schuster, S., Tyers, F., and Zeman, D. (2020c).
\newblock {U}niversal {D}ependencies v2: An evergrowing multilingual treebank
  collection.
\newblock In {\em Proceedings of the 12th Language Resources and Evaluation
  Conference}, pages 4034--4043, Marseille, France. European Language Resources
  Association.

\bibitem[Noji and Miyao(2014)]{noji-miyao-2014-left}
Noji, H. and Miyao, Y. (2014).
\newblock Left-corner transitions on dependency parsing.
\newblock In {\em Proceedings of {COLING} 2014, the 25th International
  Conference on Computational Linguistics: Technical Papers}, pages 2140--2150,
  Dublin, Ireland. Dublin City University and Association for Computational
  Linguistics.

\bibitem[Reddy {\em et~al.}(2017)]{reddy17}
Reddy, S., T{\"a}ckstr{\"o}m, O., Petrov, S., Steedman, M., and Lapata, M.
  (2017).
\newblock {Universal Semantic Parsing}.
\newblock In {\em Proceedings of the 2017 Conference on Empirical Methods in
  Natural Language Processing}, pages 89--101, Copenhagen, Denmark. Association
  for Computational Linguistics.

\bibitem[Sanada(2016)]{sanada}
Sanada, H. (2016).
\newblock The menzerath-altmann law and sentence structure.
\newblock {\em Journal of Quantitative Lingusitics}, {\bf 23}(3), 256--277.

\bibitem[Schuler {\em et~al.}(2010)]{schuler2010broad}
Schuler, W., AbdelRahman, S., Miller, T., and Schwartz, L. (2010).
\newblock {Broad-Coverage Parsing Using Human-like Memory Constraints}.
\newblock {\em Computational Linguistics}, {\bf 36}(1), 1--30.

\bibitem[Sethi and Ullman(1970)]{sethi70}
Sethi, R. and Ullman, J.~D. (1970).
\newblock {The Generation of Optimal Code for Arithmetic Expressions}.
\newblock {\em J. ACM}, {\bf 17}(4), 715--728.

\bibitem[Shannon(1948)]{shannon48}
Shannon, C.~E. (1948).
\newblock {A mathematical theory of communication}.
\newblock {\em The Bell System Technical Journal}, {\bf 27}(3), 379--423.

\bibitem[Sichel(1974)]{sichel}
Sichel, H.~S. (1974).
\newblock {On a Distribution Representing Sentence-length in written Prose}.
\newblock {\em Journal of the Royal Statistical Society: Series A}, {\bf
  137}(1), 25--34.

\bibitem[Sperling(1960)]{Sperling1960}
Sperling, G. (1960).
\newblock {The information available in brief visual presentations.}
\newblock {\em Psychological Monographs: General and Applied}, {\bf 74}(11), 1.

\bibitem[Stanley(2015)]{stanley2015catalan}
Stanley, R.~P. (2015).
\newblock {\em {Catalan numbers}}.
\newblock Cambridge University Press.

\bibitem[{Strahler}(1957)]{Strahler57}
{Strahler}, A.~N. (1957).
\newblock {Quantitative analysis of watershed geomorphology}.
\newblock {\em Eos, Transactions American Geophysical Union}, {\bf 38}(6),
  913--920.

\bibitem[Takahashi and Tanaka-Ishii(2019)]{cl19}
Takahashi, S. and Tanaka-Ishii, K. (2019).
\newblock Evaluating computational language models with scaling properties of
  natural language.
\newblock {\em Computational Lingusitics}, {\bf 45}(3), 1--33.

\bibitem[Takahira {\em et~al.}(2016)]{entropy16}
Takahira, R., Tanaka-Ishii, K., and D\c{e}bowski, {\L}. (2016).
\newblock Entropy rate estimates for natural language—a new extrapolation of
  compressed large-scale corpora.
\newblock {\em Entropy}, {\bf 18}(10), 364.

\bibitem[Tanaka-Ishii(2021)]{mal21}
Tanaka-Ishii, K. (2021).
\newblock Menzerath's law in the syntax of languages compared with random
  sentences.
\newblock {\em Entropy}.
\newblock e23060661. Selected as Editor's Choice Articles of Entropy 2022.

\bibitem[Tanaka-Ishii and Aihara(2015)]{cl15}
Tanaka-Ishii, K. and Aihara, S. (2015).
\newblock Computational constancy measures of texts -- {Yule}'s {K} and
  {R\'enyi}'s entropy.
\newblock {\em Association for Computational Linguistics}, {\bf 41}(3),
  481--502.

\bibitem[Tanaka-Ishii and Bunde(2016)]{plosone16}
Tanaka-Ishii, K. and Bunde, A. (2016).
\newblock Long-range memory in literary texts: On the universal clustering of
  the rare words.
\newblock {\em PLoS One}, {\bf 11}(11), e0164658.

\bibitem[Tanaka-Ishii and Kobayashi(2018)]{jpc18}
Tanaka-Ishii, K. and Kobayashi, T. (2018).
\newblock Taylor's law for linguistic sequences and random walk models.
\newblock {\em Journal of Physics Communications}, {\bf 2}(11), 115024.
\newblock 089401.

\bibitem[Tesni\`{e}re(1959)]{Tesniere59}
Tesni\`{e}re, L. (1959).
\newblock {\em {\'{E}lements de syntaxe structurale}}.
\newblock C. Klincksieck, Paris.

\bibitem[Tran and Miyao(2022)]{tran22}
Tran, T.-A. and Miyao, Y. (2022).
\newblock {Development of a Multilingual {CCG} Treebank via {U}niversal
  {D}ependencies Conversion}.
\newblock In {\em Proceedings of the Thirteenth Language Resources and
  Evaluation Conference}, pages 5220--5233, Marseille, France. European
  Language Resources Association.

\bibitem[Xu and Reitter(2016)]{xu2016convergence}
Xu, Y. and Reitter, D. (2016).
\newblock {Convergence of Syntactic Complexity in Conversation}.
\newblock In {\em Proceedings of the 54th Annual Meeting of the Association for
  Computational Linguistics (Volume 2: Short Papers)}, pages 443--448.

\bibitem[Yadav {\em et~al.}(2020)]{yadav20}
Yadav, H., Vaidya, A., Shukla, V., and Husain, S. (2020).
\newblock {Word Order Typology Interacts With Linguistic Complexity: A
  Cross-Linguistic Corpus Study}.
\newblock {\em Cognitive Science}, {\bf 44}(4), e12822.

\bibitem[Yang and Deng(2020)]{yang2020strongly}
Yang, K. and Deng, J. (2020).
\newblock {Strongly Incremental Constituency Parsing with Graph Neural
  Networks}.
\newblock {\em Advances in Neural Information Processing Systems}, {\bf 33},
  21687--21698.

\bibitem[Yngve(1960)]{yngve60}
Yngve, V.~H. (1960).
\newblock {A Model and an Hypothesis for Language Structure}.
\newblock {\em Proceedings of the American Philosophical Society}, {\bf
  104}(5), 444--466.

\bibitem[Yule(1968)]{yule}
Yule, U.~G. (1968).
\newblock {\em {The Statistical Study of Literary Vocabulary}}.
\newblock Archon Books.

\bibitem[Zhang(2020)]{zhang2020survey}
Zhang, M. (2020).
\newblock {A survey of syntactic-semantic parsing based on constituent and
  dependency structures}.
\newblock {\em Science China Technological Sciences}, {\bf 63}(10), 1898--1920.

\bibitem[Zipf(1949)]{zipf}
Zipf, G.~K. (1949).
\newblock {\em Human Behavior and the Principle of Least Effort : An
  Introduction to Human Ecology}.
\newblock Addison-Wesley Press.

\end{thebibliography}

\newpage

\appendix \def\thesection{\Alph{section}}

\section{Upper/Lower Limits of $n$-Node Dependency Tree}
\label{app:upperlower}
Given a dependency tree $t$, for a word in node $v$, let $f_{\max}(v)$
and $f_{\min}(v)$ be functions that obtain the maximum and minimum
Strahler numbers, respectively. These functions give the maximum and
minimum across any binarization of $t$.

Both $f_{\max}(v)$ and $f_{\min}(v)$ are calculated using the
following function $it(x, y)$:
\begin{equation*}
it(x, y) = \left\{
\begin{aligned}
&x+1& &\textrm{if } x == y\\ &\max{(x, y)}& &\textrm{otherwise}
\end{aligned}.
\right.
\end{equation*}
Note that this function is almost the same as the definition of the
Strahler number given in Section 3.1.

Using this function, the upper and lower limits are calculated in a
bottom-up manner from the leaves to the root. Let $a \Leftarrow b$
denote a computational substitution. The calculations of the
upper/lower limits, which proceed similarly, are defined below on the
left/right, respectively:

\begin{multicols}{2}
\paragraph{Upper limit} $f_{\max}(v)$ is computed as follows.

\begin{itemize}
\item For a leaf node $v$, $f_{\max}(v) \Leftarrow 1$.

\item For an inner node $v$, $f_{\max}(v)$ is calculated by the
  following four steps, where CH$(v)$ is the set of children of $v$ in
  the dependency tree.

\begin{enumerate}
\item Sort CH$(v)$ in {\em ascending} order of $f_{\max}(v')$,
  $v'\in$CH$(v)$. Let $ch_i \in $CH$(v)$, $i=1,\ldots,|$CH$(v)|$,
  denote the $i$th child in this order.
\item Set $f_{\max}(v) \Leftarrow 0$ (Initialization).\label{fmax2}
\item $f_{\max}(v) \Leftarrow it(f_{\max}(v), 1)$ (Upper limit should
  always count as 1). \label{fmax4}
\item For $ch_i \in$ CH$(v)$, repeat in order of $i$, as follows: \\
\hspace*{3mm}$f_{\max}(v) \Leftarrow it \left(f_{\max}(v),
f_{\max}(ch_i)\right)$. \label{fmax3a}
\end{enumerate}
\end{itemize}
\vspace{12pt}

\paragraph{Lower limit} $f_{\min}(v)$ is computed as follows.
\begin{itemize}
\item For a leaf node $v$, $f_{\min}(v) \Leftarrow 1$.
\vspace*{-3pt}
\item For an inner node $v$, $f_{\min}(v)$ is calculated by the
  following four steps, where CH$(v)$ is the set of children of $v$ in
  the dependency tree.
\vspace*{0pt}
\begin{enumerate}
\item Sort CH$(v)$ in {\em descending} order of $f_{\min}(v')$,
  $v'\in$ CH$(v)$. Let $ch_i \in $CH$(v)$, $i=1,\ldots |$CH$(v)|$,
  denote the $i$th child in this order.
\item Set $f_{\min}(v) \Leftarrow 0$ (Initialization). \label{fmin2}
\item For $ch_i \in$ CH$(v)$, repeat in order of $i$, as follows: \\
\hspace{3mm} $f_{\min}(v) \Leftarrow it \left(f_{\min}(v),
f_{\min}(ch_i)\right)$. \label{fmin3}
\vspace*{0pt}
\item $f_{\min}(v) \Leftarrow it(f_{\min}(v), 1)$ (Function $it$
  should be considered for node $v$). \label{fmin4}
\end{enumerate}
\end{itemize}
\vspace{0pt} {\small}
\end{multicols}

\section{Calculation of Average Upper/Lower Limits of $R(n)$}
This section briefly describes how to compute the average upper or
lower limit of $R(n)$. The method's precise details are given in
\appref{sec:average}. For a given set of trees, $R(n)$, let
$R_{n,p_{\max}}$ be {\em the total number of trees} of size $n$ with
an upper limit $p_{\max}$, and let $R_{n,p_{\min}}$ be the same with a
lower limit $p_{\min}$.

The average upper and lower limits, respectively denoted as
$R_{\max}(n)$ and $R_{\min}(n)$, are calculated as follows:
\begin{itemize}
\item Upper limit:
\begin{equation*}
 R_{\max}(n) = \frac{\sum_{p_{\max}} p_{\max}R_{n,p_{\max}}
 }{\sum_{p_{\max}} R_{n,p_{\max}}}.
\end{equation*}

\item Lower limit:
\begin{equation*}
 R_{\min}(n) = \frac{\sum_{p_{\min}}
   p_{\min}R_{n,p_{\min}}}{\sum_{p_{\min}} R_{n,p_{\min}}}.
\end{equation*}
\end{itemize}
Hence, it becomes necessary to enumerate $R_{n,p_{\max}}$ and $R_{n,
  p_{\min}}$ through dynamic programming. The details are explained in
\appref{sec:average}.

\section{Relations Among Shift-Reduce Method, Sethi-Ullman Algorithm \citep{sethi70}, and Strahler Number}
\label{sec:relations}
A computation tree is evaluated by use of the two operations of shift
and reduce. The optimal sequence of operations to minimize the number
of stack space uses is given by the Sethi-Ullman Algorithm
\citep{sethi70}. As mentioned in the main text, \citet{Ershov58}
showed that the minimum number of stack space uses equals the Strahler
number. This section clarifies the relations among the shift-reduce
method, the Sethi-Ullman algorithm \citep{sethi70}, and the Strahler
number.

\paragraph{Shift-reduce method} The shift-reduce algorithm is an algorithm for using a stack (a LIFO structure) to evaluate a computation tree. The tree is a binary tree in which each leaf node stores data and each inner node is a function that combines the values of its two children.

The shift and reduce operations are defined as follows:
\begin{itemize}
\item[]\textbf{Shift} Put a leaf from the unanalyzed part of the tree
  on the top of the stack.
\item[]\textbf{Reduce} Remove the top two elements from the stack and
  combine them into a single component according to the function in
  the corresponding inner node; then, put the result on the top of the
  stack.
\end{itemize}

\paragraph{Sethi-Ullman algorithm} The Sethi-Ullman algorithm is a depth-first evaluation method using registers for memory. The use of registers when applying a depth-first strategy to an evaluation tree is a LIFO process. The Sethi-Ullman algorithm preprocesses a given evaluation tree by annotating each node with a value, in an equivalent manner to the Strahler number, as follows:
\begin{enumerate}
\item Assign a value $x(v)$ for each node $v$.
\begin{itemize}
\item If node $v$ is a leaf, $x(v)\Leftarrow 1$.
\item If node $v$ is an inner node, let its two children be $v1,v2$.
\begin{itemize}
\item If $x(v1)=x(v2)$, then $x(v)\Leftarrow x(v1)$.
\item Otherwise, $x(v) \Leftarrow $ max$(x(v1),x(v2))$.
\end{itemize}
\end{itemize}
\item The tree evaluation is conducted depth-first, by always first
  evaluating the child $v$ with the larger $x(v)$.
\end{enumerate}
\cite{sethi70} proved that this algorithm minimizes the number of
registers used for evaluation.

\section{Calculation of Average Upper/Lower Limits of $R(n)$}
\label{sec:average}
Here, we explain the details of the method given in Appendix B to
calculate the averages of the respective upper and lower limits,
$R_{\max}(n)$ and $R_{\min}(n)$, of $R(n)$. The calculation uses
dynamic programming for the upper/lower limits by enumerating
$R_{n,p_{\max}}$ and $R_{n,p_{\min}}$, respectively. Because the
dynamic programming proceeds in the same manner for both cases, the
method is explained here for $R_{n,p}$, where $p$ indicates either
$p_{\max}$ or $p_{\min}$. Furthermore, for a node $v$, we use the
function $f(v)$ as defined in Appendix A to denote the corresponding
$f_{\max}(v)$ or $f_{\min}(v)$.

For a given tree $t$, let $\#t$ denote its size and $r(t)$ denote its
root node. For node $v$ in tree $t$, let CH$(v)$ denote the set of
children of $v$. Let $Q(t)$ be
\begin{equation*}
 Q(t) = \{f(ch) | ch \in \mathrm{CH}(r(t))\}.
\end{equation*}
Here, $Q(t)$ is a set of integers acquired as limits on the children
of $r(t)$. For each $Q(t)$, a Strahler number can be
computed. Calculation of the limits requires consideration for
$\forall t \in R(n)$, but $|R(n)|$ is a Catalan number, and the
enumeration is thus difficult. One strategy is to consider trees with
the same Strahler number as a certain type, which can be obtained from
$Q(t)$ as follows.

\begin{figure}[t]
 \centering \includegraphics[width = 0.6\linewidth]{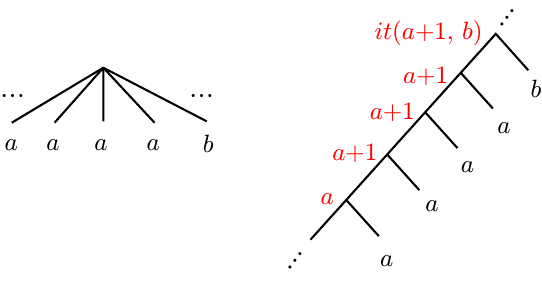}
\caption{Illustration of the equality of Strahler numbers for states
  having the same integer more than twice. The left side shows a
  dependency tree $t$ with 4 children whose Strahler number are
  same. For this tree $t$, $Q(t) = \{\cdots a,a,a,a,b \cdots\}$, and
  $Q'(t) = \{\cdots a,a,b \cdots\}$. Note that the elements of $Q(t)$
  are sorted in either ascending or descending order as described in
  Appendix A; therefore, instances of the same number will always
  appear together in the set. Instances of an integer appearing more
  than twice in $Q(t)$ can be removed, enabling reduction to $Q'(t)$,
  as follows. The right side of the figure shows the binary tree of
  $t$. The Strahler numbers of the first two nodes are $a$ and $a+1$,
  but after the second node from the left, the Strahler numbers are
  all $a+1$. This example shows how integer instances appearing more
  than twice can be eliminated. This reduction is essential to deal
  with the Catalan number size of $R(n)$.
\label{tenary}}
\end{figure}

As explained in Appendix A for the procedures to calculate the
upper/lower limits, the elements of $Q(t)$ are sorted in
ascending/descending order, and instances of the same integer element
will thus appear together. If the sorted $Q(t)$ has more than two
instances of the same integer, such as 1 in $Q(t) = \{1,1,1,1,2,3\}$,
then half those instances can be removed, yielding $\{1,1,2,3\}$,
because the resulting Strahler numbers for these sets are the
same. The reason for this is explained in the caption of
\figref{tenary}.

Hence, we define $Q'(t)$ as a reduced version of $Q(t)$ with
appropriate elimination of integers having more than two instances. A
{\em state} $q$ refers here to each such reduced $Q'(t)$. Trees with
the same state have the same upper/lower limits of the Strahler
number, denoted as $st(q)$. For the set of trees $R(n)$, the number of
trees with state $q$ and size $n$ is denoted as $S_{n,q}$.

For example, consider calculation of the upper limit for five trees in
$T = \{t_1,\ldots,t_5\}$, which all have size $n=5$. Suppose that the
trees have $Q(t)$ sets $\{2,1,2\}, \{2,1,2,2\}, \{1,1,1\}, \{1,2\},
\{2,1\}$. By sorting each $Q(t)$ in ascending order, these sets become
$\{1,2,2\}$, $\{1,2,2,2\}$, $\{1,1,1\}$, $\{1,2\}$,
$\{1,2\}$. Elimination of more than two appearances of the same
integer in each set yields $\{1,2,2\}$, $\{1,2,2\}$, $\{1,1\}$,
$\{1,2\}$, $\{1,2\}$, each of which is a state. Among these states,
$\{1,2,2\}$ and $\{1,2\}$ are redundant. Thus, the upper/lower limits
of $t1,t2$ and $t4$,$t5$ are the same, and they can be handled via the
same state type. Then, $st(q)$ can be computed for each state type
$q$, e.g., $st(\{1,2,2\}) = 3$.

Given $S_{n,q}$, $R_{n,p}$ is obviously acquired as follows:
\begin{equation}
\begin{aligned}
R_{n,p}&=\sum_{st(q)=p}S_{n, q}, \\ R_{n, p} &= 0\ \ \textrm{ if $p >
  1, n = 1$}, \\ R_{n, p} &= 1\ \ \textrm{ if $p = 1, n = 1$}.
\end{aligned}
\end{equation}

\begin{figure}[t]
 \centering \includegraphics[width = 0.6\linewidth]{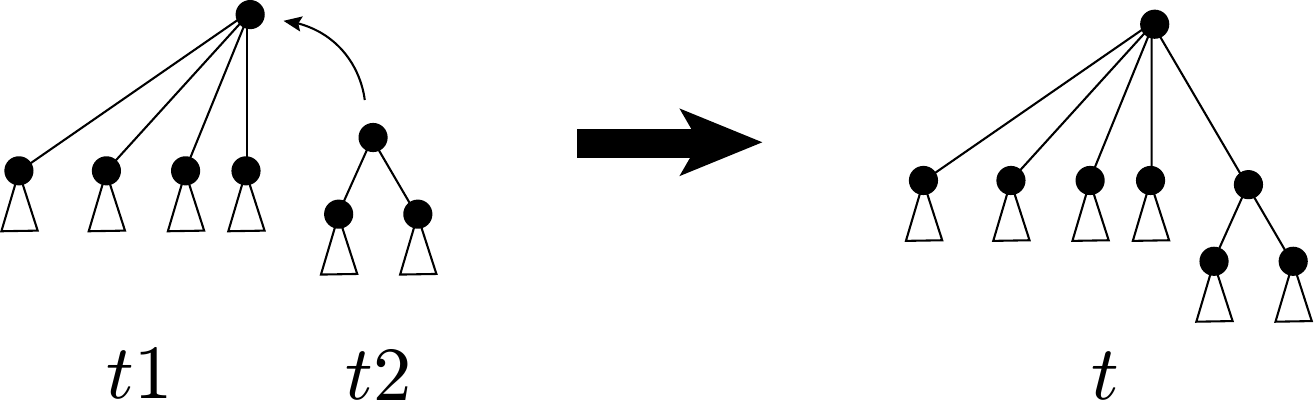}
\caption{Dynamic programming by induction across subtrees.}
 \label{subtreedp}
\end{figure}

$S_{n,q}$ is calculated inductively through dynamic programming, by
first acquiring the value for a small $n$ and then calculating the
value for a larger $n$. \figref{subtreedp} illustrates a recursive
calculation of $S_{n,q}$. The left side shows $t1$ and $t2$, which are
combined to form $t$ on the right side, where $\#t=n=\#t1+\#t2$. Let
$q1 = Q'(t1)$ and $q2 = Q'(t2)$ be the respective states of $t1$,
$t2$. We define a function $g$ as follows:
\begin{equation}
 g(q1 ,q2) \equiv q1 \oplus st(q2),
\end{equation}
where $\oplus$ denotes the following procedure.
\begin{enumerate}
\item Append $st(q2))$ as an element of $q1$;
\item Sort the elements in ascending/descending order (see Appendix
  A);
\item Eliminate integers with more than two instances (see
  \figref{tenary} here).
\end{enumerate}
Then, $S_{n,q}$ is calculated for all possible states $q1$ and $q2$ as
follows:
\begin{equation}
\begin{aligned}
S_{n, q} = \sum_{\#t1 + \#t2 = n} \sum_{q=g(t1 ,t2)} S_{\#t1, q1}
S_{\#t2, q2}\ \ \ \ \ \ \ \textrm{ for $n > 1$}.\\
\end{aligned}
\label{upperdp}
\end{equation}
To calculate $S_{n,q}$ for a tree with $\#t = n$, all trees of smaller
size must be considered, i.e., $t1$ and $t2$, where $\#t1 + \#t2 =
n$. Direct enumeration of all pairs $t1, t2$ for given $q$ and $n$
would be nontrivial because $|R(n)|$ is a Catalan number. However, as
mentioned above, the computation is always performed via a set of
states, which are far smaller in number than the set of trees, through
reduction via state types.

\newpage
\section{Details of Data}

\subsection{Datasets and numbers of sentences}
\label{sec:datanumsentences}

\begin{table}[h]
\caption{List of datasets and numbers of sentences\label{datacount2}}
  \begin{minipage}[t]{.2\textwidth}
    \begin{center}
    \footnotesize
    \begin{tabular}{p{4cm}r}
      \hline datasets & numbers of sentences \\
      \hline UD\_Akkadian-RIAO & 1907 \\ UD\_Armenian-ArmTDP & 2502 \\ UD\_Welsh-CCG & 1833
\\ UD\_Norwegian-Nynorsk & 17575 \\ UD\_Old\_East\_Slavic-TOROT &
16944 \\ UD\_English-LinES & 5243 \\ UD\_Albanian-TSA & 60
\\ UD\_French-Sequoia & 3099 \\ UD\_Hindi\_English-HIENCS & 1898
\\ UD\_Slovenian-SST & 3188 \\ UD\_Guajajara-TuDeT & 103
\\ UD\_Kurmanji-MG & 754 \\ UD\_Italian-PUD & 1000 \\ UD\_Turkish-GB &
2880 \\ UD\_Finnish-FTB & 18723 \\ UD\_Indonesian-GSD & 5593
\\ UD\_Ukrainian-IU & 7060 \\ UD\_Dutch-LassySmall & 7341
\\ UD\_Polish-PDB & 22152 \\ UD\_Turkish-Kenet & 18687
\\ UD\_Portuguese-Bosque & 9364 \\ UD\_Kazakh-KTB & 1078
\\ UD\_Italian-PoSTWITA & 6713 \\ UD\_Latin-ITTB & 26977
\\ UD\_Old\_French-SRCMF & 17678 \\ UD\_Spanish-PUD & 1000
\\ UD\_Buryat-BDT & 927 \\ UD\_Kaapor-TuDeT & 61 \\ UD\_Korean-PUD &
1000 \\ UD\_Polish-PUD & 1000 \\ UD\_Latin-Perseus & 2273
\\ UD\_Estonian-EDT & 30972 \\ UD\_Croatian-SET & 9010
\\ UD\_Gothic-PROIEL & 5401 \\ UD\_Turkish-FrameNet & 2698
\\ UD\_Latin-LLCT & 9023 \\ UD\_Swedish\_Sign\_Language-SSLC & 203
\\ UD\_Swiss\_German-UZH & 100 \\ UD\_Assyrian-AS & 57
\\ UD\_Italian-ISDT & 14167 \\ UD\_North\_Sami-Giella & 3122
\\ UD\_Norwegian-Bokmaal & 20044 \\ UD\_Naija-NSC & 9242
\\ UD\_German-LIT & 1922 \\ UD\_Latvian-LVTB & 15351
\\ UD\_Chinese-GSDSimp & 4997 \\ UD\_Tagalog-Ugnayan & 94
\\ UD\_Bambara-CRB & 1026 \\ UD\_Lithuanian-ALKSNIS & 3642 \\ \hline
\end{tabular}
    
    \end{center}
  \end{minipage}
  \begin{minipage}[t]{.3\textwidth}
    \begin{center}
    \end{center}
  \end{minipage}
  \begin{minipage}[t]{.2\textwidth}
    \begin{center}
    \footnotesize
    \begin{tabular}{p{4cm}r}
\hline datasets & numbers of sentences \\ \hline
UD\_Korean-GSD & 6339 \\ UD\_Amharic-ATT & 1074 \\ UD\_Greek-GDT &
2521 \\ UD\_German-HDT & 189928 \\ UD\_Spanish-GSD & 16013
\\ UD\_Catalan-AnCora & 16678 \\ UD\_English-PUD & 1000
\\ UD\_English-EWT & 16621 \\ UD\_Soi-AHA & 8
\\ UD\_Old\_East\_Slavic-RNC & 957 \\ UD\_Swedish-LinES & 5243
\\ UD\_Yupik-SLI & 309 \\ UD\_Russian-SynTagRus & 61889
\\ UD\_Czech-PDT & 87913 \\ UD\_Beja-NSC & 56 \\ UD\_Italian-VIT &
10087 \\ UD\_Erzya-JR & 1690 \\ UD\_Bhojpuri-BHTB & 357
\\ UD\_Polish-LFG & 17246 \\ UD\_Thai-PUD & 1000 \\ UD\_Marathi-UFAL &
466 \\ UD\_Basque-BDT & 8993 \\ UD\_Slovak-SNK & 10604 \\ UD\_Kiche-IU
& 1435 \\ UD\_Russian-Taiga & 17870 \\ UD\_Yoruba-YTB & 318
\\ UD\_Warlpiri-UFAL & 55 \\ UD\_Low\_Saxon-LSDC & 36
\\ UD\_English-Pronouns & 285 \\ UD\_Finnish-OOD & 2122
\\ UD\_French-FTB & 18535 \\ UD\_Tamil-TTB & 600 \\ UD\_Spanish-AnCora
& 17680 \\ UD\_Maltese-MUDT & 2074 \\ UD\_Finnish-TDT & 15136
\\ UD\_Indonesian-CSUI & 1030 \\ UD\_Ancient\_Greek-Perseus & 13919
\\ UD\_Icelandic-IcePaHC & 44029 \\ UD\_Mbya\_Guarani-Thomas & 98
\\ UD\_Urdu-UDTB & 5130 \\ UD\_Romanian-RRT & 9524 \\ UD\_Tagalog-TRG
& 128 \\ UD\_Akkadian-PISANDUB & 101 \\ UD\_Latin-PROIEL & 18411
\\ UD\_Persian-PerDT & 29107 \\ UD\_Apurina-UFPA & 88
\\ UD\_Indonesian-PUD & 1000 \\ UD\_Japanese-Modern & 822 \\ \hline
\end{tabular}
    \end{center}
  \end{minipage}
  \label{datacount}
\end{table}

\begin{table}[h]
  \begin{minipage}[t]{.2\textwidth}
    \begin{center}
    \footnotesize
    \begin{tabular}{p{4cm}r}
\hline datasets & numbers of sentences \\ \hline

UD\_Galician-CTG & 3993 \\ UD\_Chinese-HK & 1004 \\ UD\_Vietnamese-VTB
& 3000 \\ UD\_Hindi-HDTB & 16647 \\ UD\_Classical\_Chinese-Kyoto &
55514 \\ UD\_Norwegian-NynorskLIA & 5250 \\ UD\_Komi\_Permyak-UH & 81
\\ UD\_Chinese-CFL & 451 \\ UD\_Faroese-FarPaHC & 1621
\\ UD\_Sanskrit-Vedic & 3997 \\ UD\_Turkish-IMST & 5635
\\ UD\_Lithuanian-HSE & 263 \\ UD\_Slovenian-SSJ & 8000
\\ UD\_Livvi-KKPP & 125 \\ UD\_Swedish-Talbanken & 6026
\\ UD\_Arabic-NYUAD & 19738 \\ UD\_English-GUMReddit & 895
\\ UD\_Italian-ParTUT & 2090 \\ UD\_Czech-FicTree & 12760
\\ UD\_Wolof-WTB & 2107 \\ UD\_Bulgarian-BTB & 11138
\\ UD\_Russian-PUD & 1000 \\ UD\_Portuguese-PUD & 1000
\\ UD\_Romanian-ArT & 50 \\ UD\_Akuntsu-TuDeT & 101
\\ UD\_Makurap-TuDeT & 31 \\ UD\_Turkish-Penn & 9557
\\ UD\_Kangri-KDTB & 288 \\ UD\_Italian-Valico & 398
\\ UD\_Japanese-BCCWJ & 57109 \\ UD\_Italian-TWITTIRO & 1424
\\ UD\_Ancient\_Greek-PROIEL & 17080 \\ UD\_Breton-KEB & 888
\\ UD\_Czech-CAC & 24709 \\ UD\_Japanese-GSD & 8100 \\ UD\_Telugu-MTG
& 1328 \\ UD\_Korean-Kaist & 27363 \\ UD\_Cantonese-HK & 1004
\\ UD\_Turkish\_German-SAGT & 2184 \\ UD\_English-GUM & 7402
\\ UD\_Chinese-PUD & 1000 \\ UD\_Romanian-Nonstandard & 26225
\\ UD\_German-GSD & 15590 \\ UD\_Old\_Church\_Slavonic-PROIEL & 6338
\\ UD\_Karelian-KKPP & 228 \\ UD\_French-PUD & 1000
\\ UD\_Upper\_Sorbian-UFAL & 646 \\ UD\_Danish-DDT & 5512
\\ UD\_Icelandic-PUD & 1000 \\ UD\_Galician-TreeGal & 1000
\\ UD\_English-ESL & 5124 \\ UD\_Hindi-PUD & 1000
\\ UD\_South\_Levantine\_Arabic-MADAR & 100 \\ \hline
\end{tabular}
    
    \end{center}
  \end{minipage}
  \begin{minipage}[t]{.3\textwidth}
    \begin{center}
    \end{center}
  \end{minipage}
  \begin{minipage}[t]{.2\textwidth}
    \begin{center}
    \footnotesize
    \begin{tabular}{p{4cm}r}
\hline datasets & numbers of sentences \\ \hline UD\_French-ParTUT & 1020
\\ UD\_Icelandic-Modern & 6928 \\ UD\_Hungarian-Szeged & 1800
\\ UD\_Komi\_Zyrian-Lattice & 658 \\ UD\_French-Spoken & 2837
\\ UD\_Sanskrit-UFAL & 230 \\ UD\_Tamil-MWTT & 534 \\ UD\_Turkish-PUD
& 1000 \\ UD\_Irish-IDT & 4910 \\ UD\_Faroese-OFT & 1208
\\ UD\_Nayini-AHA & 10 \\ UD\_Frisian\_Dutch-Fame & 400
\\ UD\_Munduruku-TuDeT & 109 \\ UD\_Manx-Cadhan & 2319
\\ UD\_Skolt\_Sami-Giellagas & 178 \\ UD\_Finnish-PUD & 1000
\\ UD\_Mbya\_Guarani-Dooley & 1046 \\ UD\_Persian-Seraji & 5997
\\ UD\_Afrikaans-AfriBooms & 1934 \\ UD\_Japanese-PUD & 1000
\\ UD\_Czech-CLTT & 1125 \\ UD\_French-GSD & 16341 \\ UD\_German-PUD &
1000 \\ UD\_Irish-TwittIrish & 866 \\ UD\_Dutch-Alpino & 13603
\\ UD\_Swedish-PUD & 1000 \\ UD\_Chinese-GSD & 4997
\\ UD\_Old\_Turkish-Tonqq & 18 \\ UD\_Romanian-SiMoNERo & 4681
\\ UD\_Arabic-PUD & 1000 \\ UD\_Tupinamba-TuDeT & 210
\\ UD\_Belarusian-HSE & 25231 \\ UD\_Coptic-Scriptorium & 1873
\\ UD\_French-FQB & 2289 \\ UD\_Komi\_Zyrian-IKDP & 214
\\ UD\_Serbian-SET & 4384 \\ UD\_Turkish-Tourism & 19749
\\ UD\_Estonian-EWT & 5536 \\ UD\_Moksha-JR & 313
\\ UD\_Western\_Armenian-ArmTDP & 1780 \\ UD\_Czech-PUD & 1000
\\ UD\_Scottish\_Gaelic-ARCOSG & 3798 \\ UD\_Portuguese-GSD & 12078
\\ UD\_Russian-GSD & 5030 \\ UD\_Khunsari-AHA & 10 \\ UD\_Turkish-BOUN
& 9761 \\ UD\_Arabic-PADT & 7664 \\ UD\_Hebrew-HTB & 6216
\\ UD\_English-ParTUT & 2090 \\ UD\_Uyghur-UDT & 3456
\\ UD\_Latin-UDante & 1721 \\ UD\_Chukchi-HSE & 1004 \\ \hline
\end{tabular}
    
    \end{center}
  \end{minipage}

\end{table} 

\newpage

\subsection{Averages and standard deviations for Binary1}

\begin{table}[h]
\caption{List of Binary1 results for each dataset\label{datacount11}}
  \begin{minipage}[t]{.2\textwidth}
    \begin{center}
    \footnotesize
    \begin{tabular}{lr}
\hline datasets & Binary1 \\ \hline UD\_Korean-Kaist & 3.19$\pm$0.57
\\ UD\_Faroese-OFT & 2.75$\pm$0.50 \\ UD\_Latin-Perseus &
3.06$\pm$0.66 \\ UD\_Finnish-TDT & 3.02$\pm$0.64 \\ UD\_Swedish-LinES
& 3.21$\pm$0.68 \\ UD\_Lithuanian-HSE & 3.39$\pm$0.55
\\ UD\_Kaapor-TuDeT & 2.26$\pm$0.44 \\ UD\_Belarusian-HSE &
2.90$\pm$0.69 \\ UD\_South\_Levantine\_Arabic-MADAR & 2.77$\pm$0.51
\\ UD\_Komi\_Zyrian-IKDP & 2.90$\pm$0.63
\\ UD\_Swedish\_Sign\_Language-SSLC & 2.69$\pm$0.70
\\ UD\_Norwegian-NynorskLIA & 2.73$\pm$0.70 \\ UD\_Russian-Taiga &
2.83$\pm$0.69 \\ UD\_Tamil-MWTT & 2.19$\pm$0.40 \\ UD\_Indonesian-CSUI
& 3.71$\pm$0.56 \\ UD\_Italian-PUD & 3.68$\pm$0.54 \\ UD\_Swedish-PUD
& 3.43$\pm$0.57 \\ UD\_English-ParTUT & 3.51$\pm$0.62
\\ UD\_Hindi\_English-HIENCS & 3.16$\pm$0.47 \\ UD\_French-ParTUT &
3.64$\pm$0.66 \\ UD\_Gothic-PROIEL & 2.90$\pm$0.69 \\ UD\_Naija-NSC &
2.92$\pm$0.68 \\ UD\_Latin-PROIEL & 2.91$\pm$0.71
\\ UD\_Hungarian-Szeged & 3.55$\pm$0.61 \\ UD\_English-GUMReddit &
3.17$\pm$0.75 \\ UD\_Akkadian-RIAO & 3.03$\pm$0.64 \\ UD\_Chinese-HK &
2.70$\pm$0.64 \\ UD\_Kazakh-KTB & 2.85$\pm$0.59 \\ UD\_German-PUD &
3.51$\pm$0.56 \\ UD\_Serbian-SET & 3.51$\pm$0.59
\\ UD\_Armenian-ArmTDP & 3.38$\pm$0.73 \\ UD\_Beja-NSC & 3.23$\pm$0.60
\\ UD\_Basque-BDT & 3.15$\pm$0.59 \\ UD\_Croatian-SET & 3.51$\pm$0.60
\\ UD\_Romanian-ArT & 3.08$\pm$0.48 \\ UD\_Icelandic-IcePaHC &
3.55$\pm$0.69 \\ UD\_Polish-PDB & 3.22$\pm$0.62
\\ UD\_Afrikaans-AfriBooms & 3.57$\pm$0.57 \\ UD\_Komi\_Permyak-UH &
2.88$\pm$0.57 \\ UD\_Czech-FicTree & 2.91$\pm$0.73
\\ UD\_Old\_East\_Slavic-RNC & 3.61$\pm$0.81 \\ UD\_Turkish-PUD &
3.34$\pm$0.54 \\ UD\_Thai-PUD & 3.65$\pm$0.56 \\ UD\_Spanish-AnCora &
3.77$\pm$0.68 \\ UD\_Persian-Seraji & 3.51$\pm$0.67 \\ UD\_Irish-IDT &
3.59$\pm$0.68 \\ UD\_Finnish-FTB & 2.80$\pm$0.60
\\ UD\_Munduruku-TuDeT & 2.50$\pm$0.50 \\ UD\_Chinese-PUD &
3.56$\pm$0.55 \\ UD\_Akuntsu-TuDeT & 2.25$\pm$0.43
\\ UD\_Norwegian-Nynorsk & 3.14$\pm$0.70 \\ UD\_Portuguese-Bosque &
3.51$\pm$0.75 \\ \hline
\end{tabular}
    
    \end{center}
  \end{minipage}
  \begin{minipage}[t]{.3\textwidth}
    \begin{center}
    \end{center}
  \end{minipage}
  \begin{minipage}[t]{.2\textwidth}
    \begin{center}
    \footnotesize
    \begin{tabular}{lr}
\hline datasets & Binary1 \\ \hline

UD\_Chukchi-HSE & 2.46$\pm$0.54 \\ UD\_Buryat-BDT & 2.95$\pm$0.63
\\ UD\_Manx-Cadhan & 3.01$\pm$0.44 \\ UD\_Dutch-Alpino & 3.16$\pm$0.66
\\ UD\_Skolt\_Sami-Giellagas & 2.96$\pm$0.56 \\ UD\_Turkish-FrameNet &
2.78$\pm$0.48 \\ UD\_Japanese-BCCWJ & 3.25$\pm$0.80
\\ UD\_Portuguese-GSD & 3.67$\pm$0.61 \\ UD\_Galician-TreeGal &
3.55$\pm$0.76 \\ UD\_Akkadian-PISANDUB & 3.46$\pm$0.67
\\ UD\_Nayini-AHA & 2.80$\pm$0.40 \\ UD\_French-FTB & 3.77$\pm$0.70
\\ UD\_Korean-GSD & 3.06$\pm$0.69 \\ UD\_Japanese-Modern &
3.22$\pm$0.73 \\ UD\_Wolof-WTB & 3.34$\pm$0.60 \\ UD\_Japanese-GSD &
3.48$\pm$0.64 \\ UD\_Classical\_Chinese-Kyoto & 2.47$\pm$0.62
\\ UD\_French-GSD & 3.66$\pm$0.60 \\ UD\_Slovenian-SSJ & 3.25$\pm$0.63
\\ UD\_Hebrew-HTB & 3.60$\pm$0.64 \\ UD\_Kangri-KDTB & 2.78$\pm$0.48
\\ UD\_Finnish-OOD & 2.64$\pm$0.71 \\ UD\_Arabic-PUD & 3.53$\pm$0.56
\\ UD\_Low\_Saxon-LSDC & 3.75$\pm$0.55 \\ UD\_Spanish-GSD &
3.76$\pm$0.60 \\ UD\_Old\_East\_Slavic-TOROT & 2.81$\pm$0.65
\\ UD\_Welsh-CCG & 3.42$\pm$0.64 \\ UD\_Moksha-JR & 2.85$\pm$0.48
\\ UD\_Danish-DDT & 3.20$\pm$0.73 \\ UD\_Catalan-AnCora &
3.85$\pm$0.63 \\ UD\_Chinese-CFL & 3.22$\pm$0.65 \\ UD\_Russian-PUD &
3.43$\pm$0.56 \\ UD\_Ancient\_Greek-PROIEL & 3.07$\pm$0.72
\\ UD\_Ancient\_Greek-Perseus & 3.15$\pm$0.65 \\ UD\_Czech-PUD &
3.35$\pm$0.56 \\ UD\_Hindi-HDTB & 3.44$\pm$0.54 \\ UD\_English-LinES &
3.25$\pm$0.67 \\ UD\_Bulgarian-BTB & 3.04$\pm$0.64 \\ UD\_Galician-CTG
& 3.98$\pm$0.39 \\ UD\_Urdu-UDTB & 3.63$\pm$0.56 \\ UD\_Indonesian-PUD
& 3.44$\pm$0.56 \\ UD\_Turkish-GB & 2.48$\pm$0.55
\\ UD\_Coptic-Scriptorium & 3.62$\pm$0.64 \\ UD\_Old\_French-SRCMF &
2.85$\pm$0.62 \\ UD\_Polish-PUD & 3.39$\pm$0.55
\\ UD\_Italian-TWITTIRO & 3.60$\pm$0.52 \\ UD\_Italian-Valico &
3.23$\pm$0.69 \\ UD\_Estonian-EWT & 2.87$\pm$0.76 \\ UD\_Turkish-BOUN
& 3.00$\pm$0.69 \\ UD\_Upper\_Sorbian-UFAL & 3.28$\pm$0.62
\\ UD\_Tagalog-Ugnayan & 3.16$\pm$0.53 \\ UD\_Bhojpuri-BHTB &
3.31$\pm$0.61 \\ \hline
\end{tabular}
    
    \end{center}
  \end{minipage}

\end{table}

\begin{table}[h]
\caption{List of Binary1 results for each dataset \label{datacount12}}
  \begin{minipage}[t]{.2\textwidth}
    \begin{center}
    \footnotesize
    \begin{tabular}{lr}
\hline datasets & Binary1 \\ \hline

UD\_German-GSD & 3.39$\pm$0.59 \\ UD\_Sanskrit-Vedic & 2.60$\pm$0.60
\\ UD\_Maltese-MUDT & 3.35$\pm$0.74 \\ UD\_Arabic-NYUAD &
3.82$\pm$0.74 \\ UD\_Finnish-PUD & 3.25$\pm$0.55
\\ UD\_Guajajara-TuDeT & 2.66$\pm$0.47 \\ UD\_Chinese-GSD &
3.67$\pm$0.56 \\ UD\_Tupinamba-TuDeT & 2.64$\pm$0.60
\\ UD\_Slovenian-SST & 2.45$\pm$0.97 \\ UD\_Scottish\_Gaelic-ARCOSG &
3.29$\pm$0.86 \\ UD\_Dutch-LassySmall & 2.90$\pm$0.90
\\ UD\_Italian-ParTUT & 3.66$\pm$0.63 \\ UD\_English-ESL &
3.32$\pm$0.60 \\ UD\_Western\_Armenian-ArmTDP & 3.38$\pm$0.72
\\ UD\_Kurmanji-MG & 3.16$\pm$0.52 \\ UD\_Cantonese-HK & 2.91$\pm$0.73
\\ UD\_Swedish-Talbanken & 3.19$\pm$0.70
\\ UD\_Old\_Church\_Slavonic-PROIEL & 2.82$\pm$0.69 \\ UD\_Telugu-MTG
& 2.28$\pm$0.46 \\ UD\_Latin-LLCT & 3.34$\pm$0.88
\\ UD\_Old\_Turkish-Tonqq & 3.11$\pm$0.46 \\ UD\_Albanian-TSA &
3.30$\pm$0.49 \\ UD\_Czech-CLTT & 3.33$\pm$1.16
\\ UD\_Romanian-Nonstandard & 3.46$\pm$0.66 \\ UD\_Arabic-PADT &
3.82$\pm$0.79 \\ UD\_Italian-VIT & 3.63$\pm$0.74 \\ UD\_Czech-CAC &
3.37$\pm$0.66 \\ UD\_Greek-GDT & 3.58$\pm$0.72 \\ UD\_Marathi-UFAL &
2.70$\pm$0.57 \\ UD\_Turkish-Kenet & 2.91$\pm$0.56
\\ UD\_Frisian\_Dutch-Fame & 2.86$\pm$0.50 \\ UD\_Latin-UDante &
3.75$\pm$0.66 \\ UD\_Icelandic-Modern & 3.45$\pm$0.68
\\ UD\_Norwegian-Bokmaal & 3.06$\pm$0.68 \\ UD\_Makurap-TuDeT &
2.35$\pm$0.48 \\ UD\_Karelian-KKPP & 3.11$\pm$0.56
\\ UD\_Faroese-FarPaHC & 3.74$\pm$0.60 \\ UD\_Livvi-KKPP &
3.07$\pm$0.54 \\ UD\_Mbya\_Guarani-Dooley & 3.01$\pm$0.50
\\ UD\_Polish-LFG & 2.65$\pm$0.55 \\ UD\_Bambara-CRB & 2.96$\pm$0.63
\\ UD\_Russian-GSD & 3.38$\pm$0.61 \\ UD\_Mbya\_Guarani-Thomas &
3.00$\pm$0.65 \\ UD\_Italian-ISDT & 3.44$\pm$0.70 \\ UD\_Apurina-UFPA
& 2.62$\pm$0.57 \\ UD\_Turkish\_German-SAGT & 3.20$\pm$0.59
\\ UD\_Icelandic-PUD & 3.44$\pm$0.56 \\ UD\_Chinese-GSDSimp &
3.67$\pm$0.56 \\ UD\_German-LIT & 3.48$\pm$0.63 \\ UD\_Soi-AHA &
2.75$\pm$0.43 \\

\hline
\end{tabular}
    
    \end{center}
  \end{minipage}
  \begin{minipage}[t]{.3\textwidth}
    \begin{center}
    \end{center}
  \end{minipage}
  \begin{minipage}[t]{.2\textwidth}
    \begin{center}
    \footnotesize
    \begin{tabular}{lr}
\hline datasets & Binary1 \\ \hline UD\_Korean-Kaist & 3.26 $\pm$ 0.54
\\ UD\_Faroese-OFT & 2.67 $\pm$ 0.50 \\ UD\_Latin-Perseus & 3.08 $\pm$
0.61 \\ UD\_Finnish-TDT & 2.93 $\pm$ 0.61 \\ UD\_Swedish-LinES & 3.09
$\pm$ 0.65 \\ UD\_Lithuanian-HSE & 3.21 $\pm$ 0.49 \\ UD\_Kaapor-TuDeT
& 2.26 $\pm$ 0.44 \\ UD\_Belarusian-HSE & 2.78 $\pm$ 0.63
\\ UD\_South\_Levantine\_Arabic-MADAR & 2.74 $\pm$ 0.50
\\ UD\_Komi\_Zyrian-IKDP & 2.84 $\pm$ 0.61
\\ UD\_Swedish\_Sign\_Language-SSLC & 2.65 $\pm$ 0.67
\\ UD\_Norwegian-NynorskLIA & 2.70 $\pm$ 0.68 \\ UD\_Russian-Taiga &
2.72 $\pm$ 0.64 \\ UD\_Tamil-MWTT & 2.45 $\pm$ 0.50
\\ UD\_Indonesian-CSUI & 3.63 $\pm$ 0.57 \\ UD\_Italian-PUD & 3.33
$\pm$ 0.55 \\ UD\_Swedish-PUD & 3.23 $\pm$ 0.52 \\ UD\_English-ParTUT
& 3.24 $\pm$ 0.57 \\ UD\_Hindi\_English-HIENCS & 3.15 $\pm$ 0.45
\\ UD\_French-ParTUT & 3.29 $\pm$ 0.61 \\ UD\_Gothic-PROIEL & 2.84
$\pm$ 0.67 \\ UD\_Naija-NSC & 2.75 $\pm$ 0.62 \\ UD\_Latin-PROIEL &
2.86 $\pm$ 0.69 \\ UD\_Hungarian-Szeged & 3.45 $\pm$ 0.59
\\ UD\_English-GUMReddit & 2.96 $\pm$ 0.70 \\ UD\_Akkadian-RIAO & 3.00
$\pm$ 0.61 \\ UD\_Chinese-HK & 2.68 $\pm$ 0.64 \\ UD\_Kazakh-KTB &
3.03 $\pm$ 0.52 \\ UD\_German-PUD & 3.32 $\pm$ 0.53 \\ UD\_Serbian-SET
& 3.32 $\pm$ 0.58 \\ UD\_Armenian-ArmTDP & 3.30 $\pm$ 0.69
\\ UD\_Beja-NSC & 3.48 $\pm$ 0.65 \\ UD\_Basque-BDT & 3.15 $\pm$ 0.56
\\ UD\_Croatian-SET & 3.30 $\pm$ 0.58 \\ UD\_Romanian-ArT & 2.96 $\pm$
0.49 \\ UD\_Icelandic-IcePaHC & 3.45 $\pm$ 0.69 \\ UD\_Polish-PDB &
3.12 $\pm$ 0.59 \\ UD\_Afrikaans-AfriBooms & 3.39 $\pm$ 0.53
\\ UD\_Komi\_Permyak-UH & 2.78 $\pm$ 0.52 \\ UD\_Czech-FicTree & 2.84
$\pm$ 0.67 \\ UD\_Old\_East\_Slavic-RNC & 3.49 $\pm$ 0.76
\\ UD\_Turkish-PUD & 3.49 $\pm$ 0.53 \\ UD\_Thai-PUD & 3.61 $\pm$ 0.56
\\ UD\_Spanish-AnCora & 3.51 $\pm$ 0.64 \\ UD\_Persian-Seraji & 3.39
$\pm$ 0.60 \\ UD\_Irish-IDT & 3.52 $\pm$ 0.67 \\ UD\_Finnish-FTB &
2.74 $\pm$ 0.58 \\ UD\_Munduruku-TuDeT & 2.41 $\pm$ 0.49
\\ UD\_Chinese-PUD & 3.60 $\pm$ 0.56 \\ UD\_Akuntsu-TuDeT & 2.26 $\pm$
0.46 \\ \hline
\end{tabular}
    
    \end{center}
  \end{minipage}

\end{table} 

\clearpage

\subsection{Averages and standard deviations for Binary2}

\begin{table}[h]
\caption{List of Binary2 results for each dataset \label{datacount21}}
  \begin{minipage}[t]{.2\textwidth}
    \begin{center}
    \footnotesize
    \begin{tabular}{lr}
\hline datasets & Binary2 \\ \hline UD\_Korean-Kaist & 3.19$\pm$0.57
\\ UD\_Faroese-OFT & 2.75$\pm$0.50 \\ UD\_Latin-Perseus &
3.06$\pm$0.66 \\ UD\_Finnish-TDT & 3.02$\pm$0.64 \\ UD\_Swedish-LinES
& 3.21$\pm$0.68 \\ UD\_Lithuanian-HSE & 3.39$\pm$0.55
\\ UD\_Kaapor-TuDeT & 2.26$\pm$0.44 \\ UD\_Belarusian-HSE &
2.90$\pm$0.69 \\ UD\_South\_Levantine\_Arabic-MADAR & 2.77$\pm$0.51
\\ UD\_Komi\_Zyrian-IKDP & 2.90$\pm$0.63
\\ UD\_Swedish\_Sign\_Language-SSLC & 2.69$\pm$0.70
\\ UD\_Norwegian-NynorskLIA & 2.73$\pm$0.70 \\ UD\_Russian-Taiga &
2.83$\pm$0.69 \\ UD\_Tamil-MWTT & 2.19$\pm$0.40 \\ UD\_Indonesian-CSUI
& 3.71$\pm$0.56 \\ UD\_Italian-PUD & 3.68$\pm$0.54 \\ UD\_Swedish-PUD
& 3.43$\pm$0.57 \\ UD\_English-ParTUT & 3.51$\pm$0.62
\\ UD\_Hindi\_English-HIENCS & 3.16$\pm$0.47 \\ UD\_French-ParTUT &
3.64$\pm$0.66 \\ UD\_Gothic-PROIEL & 2.90$\pm$0.69 \\ UD\_Naija-NSC &
2.92$\pm$0.68 \\ UD\_Latin-PROIEL & 2.91$\pm$0.71
\\ UD\_Hungarian-Szeged & 3.55$\pm$0.61 \\ UD\_English-GUMReddit &
3.17$\pm$0.75 \\ UD\_Akkadian-RIAO & 3.03$\pm$0.64 \\ UD\_Chinese-HK &
2.70$\pm$0.64 \\ UD\_Kazakh-KTB & 2.85$\pm$0.59 \\ UD\_German-PUD &
3.51$\pm$0.56 \\ UD\_Serbian-SET & 3.51$\pm$0.59
\\ UD\_Armenian-ArmTDP & 3.38$\pm$0.73 \\ UD\_Beja-NSC & 3.23$\pm$0.60
\\ UD\_Basque-BDT & 3.15$\pm$0.59 \\ UD\_Croatian-SET & 3.51$\pm$0.60
\\ UD\_Romanian-ArT & 3.08$\pm$0.48 \\ UD\_Icelandic-IcePaHC &
3.55$\pm$0.69 \\ UD\_Polish-PDB & 3.22$\pm$0.62
\\ UD\_Afrikaans-AfriBooms & 3.57$\pm$0.57 \\ UD\_Komi\_Permyak-UH &
2.88$\pm$0.57 \\ UD\_Czech-FicTree & 2.91$\pm$0.73
\\ UD\_Old\_East\_Slavic-RNC & 3.61$\pm$0.81 \\ UD\_Turkish-PUD &
3.34$\pm$0.54 \\ UD\_Thai-PUD & 3.65$\pm$0.56 \\ UD\_Spanish-AnCora &
3.77$\pm$0.68 \\ UD\_Persian-Seraji & 3.51$\pm$0.67 \\ UD\_Irish-IDT &
3.59$\pm$0.68 \\ UD\_Finnish-FTB & 2.80$\pm$0.60
\\ UD\_Munduruku-TuDeT & 2.50$\pm$0.50 \\ UD\_Chinese-PUD &
3.56$\pm$0.55 \\ \hline
\end{tabular}
    
    \end{center}
  \end{minipage}
  \begin{minipage}[t]{.3\textwidth}
    \begin{center}
    \end{center}
  \end{minipage}
  \begin{minipage}[t]{.2\textwidth}
    \begin{center}
    \footnotesize
    \begin{tabular}{lr}
\hline datasets & Binary2 \\ \hline

UD\_Akuntsu-TuDeT & 2.25$\pm$0.43 \\ UD\_Norwegian-Nynorsk &
3.14$\pm$0.70 \\ UD\_Portuguese-Bosque & 3.51$\pm$0.75
\\ UD\_Norwegian-Nynorsk & 2.99 $\pm$ 0.66 \\ UD\_Portuguese-Bosque &
3.25 $\pm$ 0.68 \\ UD\_Chukchi-HSE & 2.47 $\pm$ 0.53 \\ UD\_Buryat-BDT
& 3.05 $\pm$ 0.56 \\ UD\_Manx-Cadhan & 2.99 $\pm$ 0.42
\\ UD\_Dutch-Alpino & 3.01 $\pm$ 0.59 \\ UD\_Skolt\_Sami-Giellagas &
2.81 $\pm$ 0.58 \\ UD\_Turkish-FrameNet & 2.95 $\pm$ 0.45
\\ UD\_Japanese-BCCWJ & 3.51 $\pm$ 0.88 \\ UD\_Portuguese-GSD & 3.37
$\pm$ 0.58 \\ UD\_Galician-TreeGal & 3.26 $\pm$ 0.68
\\ UD\_Akkadian-PISANDUB & 3.42 $\pm$ 0.66 \\ UD\_Nayini-AHA & 3.00
$\pm$ 0.00 \\ UD\_French-FTB & 3.49 $\pm$ 0.66 \\ UD\_Korean-GSD &
3.14 $\pm$ 0.68 \\ UD\_Japanese-Modern & 3.50 $\pm$ 0.79
\\ UD\_Wolof-WTB & 3.22 $\pm$ 0.59 \\ UD\_Japanese-GSD & 3.77 $\pm$
0.66 \\ UD\_Classical\_Chinese-Kyoto & 2.43 $\pm$ 0.61
\\ UD\_French-GSD & 3.32 $\pm$ 0.56 \\ UD\_Slovenian-SSJ & 3.02 $\pm$
0.56 \\ UD\_Hebrew-HTB & 3.50 $\pm$ 0.62 \\ UD\_Kangri-KDTB & 2.94
$\pm$ 0.45 \\ UD\_Finnish-OOD & 2.57 $\pm$ 0.68 \\ UD\_Arabic-PUD &
3.48 $\pm$ 0.57 \\ UD\_Low\_Saxon-LSDC & 3.56 $\pm$ 0.55
\\ UD\_Spanish-GSD & 3.39 $\pm$ 0.58 \\ UD\_Old\_East\_Slavic-TOROT &
2.78 $\pm$ 0.64 \\ UD\_Welsh-CCG & 3.25 $\pm$ 0.61 \\ UD\_Moksha-JR &
2.80 $\pm$ 0.46 \\ UD\_Danish-DDT & 3.08 $\pm$ 0.68
\\ UD\_Catalan-AnCora & 3.54 $\pm$ 0.61 \\ UD\_Chinese-CFL & 3.20
$\pm$ 0.59 \\ UD\_Russian-PUD & 3.22 $\pm$ 0.52
\\ UD\_Ancient\_Greek-PROIEL & 2.96 $\pm$ 0.68
\\ UD\_Ancient\_Greek-Perseus & 3.11 $\pm$ 0.61 \\ UD\_Czech-PUD &
3.15 $\pm$ 0.50 \\ UD\_Hindi-HDTB & 3.58 $\pm$ 0.54
\\ UD\_English-LinES & 3.05 $\pm$ 0.62 \\ UD\_Bulgarian-BTB & 2.89
$\pm$ 0.58 \\ UD\_Galician-CTG & 3.64 $\pm$ 0.49 \\ UD\_Urdu-UDTB &
3.72 $\pm$ 0.54 \\ UD\_Indonesian-PUD & 3.35 $\pm$ 0.54
\\ UD\_Turkish-GB & 2.70 $\pm$ 0.54 \\ UD\_Coptic-Scriptorium & 3.46
$\pm$ 0.62 \\ UD\_Old\_French-SRCMF & 2.70 $\pm$ 0.57 \\ \hline
\end{tabular}
    
    \end{center}
  \end{minipage}

\end{table}

\begin{table}[h]
\caption{List of Binary2 results for each dataset \label{datacount22}}
  \begin{minipage}[t]{.2\textwidth}
    \begin{center}
    \footnotesize
    \begin{tabular}{lr}
\hline datasets & Binary2 \\ \hline

UD\_Polish-PUD & 3.28 $\pm$ 0.55 \\ UD\_Italian-TWITTIRO & 3.27 $\pm$
0.48 \\ UD\_Italian-Valico & 2.98 $\pm$ 0.58 \\ UD\_Estonian-EWT &
2.80 $\pm$ 0.71 \\ UD\_Turkish-BOUN & 3.12 $\pm$ 0.65
\\ UD\_Upper\_Sorbian-UFAL & 3.17 $\pm$ 0.54 \\ UD\_Tagalog-Ugnayan &
3.07 $\pm$ 0.53 \\ UD\_Bhojpuri-BHTB & 3.43 $\pm$ 0.63
\\ UD\_German-GSD & 3.20 $\pm$ 0.52 \\ UD\_Sanskrit-Vedic & 2.57 $\pm$
0.58 \\ UD\_Maltese-MUDT & 3.16 $\pm$ 0.69 \\ UD\_Arabic-NYUAD & 3.80
$\pm$ 0.73 \\ UD\_Finnish-PUD & 3.15 $\pm$ 0.51 \\ UD\_Guajajara-TuDeT
& 2.70 $\pm$ 0.48 \\ UD\_Chinese-GSD & 3.64 $\pm$ 0.54
\\ UD\_Tupinamba-TuDeT & 2.64 $\pm$ 0.63 \\ UD\_Slovenian-SST & 2.34
$\pm$ 0.90 \\ UD\_Scottish\_Gaelic-ARCOSG & 3.23 $\pm$ 0.84
\\ UD\_Dutch-LassySmall & 2.75 $\pm$ 0.81 \\ UD\_Italian-ParTUT & 3.31
$\pm$ 0.58 \\ UD\_English-ESL & 3.05 $\pm$ 0.55
\\ UD\_Western\_Armenian-ArmTDP & 3.27 $\pm$ 0.67 \\ UD\_Kurmanji-MG &
3.14 $\pm$ 0.46 \\ UD\_Cantonese-HK & 2.90 $\pm$ 0.72
\\ UD\_Swedish-Talbanken & 3.05 $\pm$ 0.66
\\ UD\_Old\_Church\_Slavonic-PROIEL & 2.79 $\pm$ 0.67
\\ UD\_Telugu-MTG & 2.48 $\pm$ 0.51 \\ UD\_Latin-LLCT & 3.40 $\pm$
0.78 \\ UD\_Old\_Turkish-Tonqq & 3.22 $\pm$ 0.42 \\ UD\_Albanian-TSA &
3.13 $\pm$ 0.39 \\ UD\_Czech-CLTT & 3.12 $\pm$ 1.07
\\ UD\_Romanian-Nonstandard & 3.29 $\pm$ 0.62 \\ UD\_Arabic-PADT &
3.78 $\pm$ 0.78 \\ UD\_Italian-VIT & 3.30 $\pm$ 0.67 \\ UD\_Czech-CAC
& 3.17 $\pm$ 0.59 \\ UD\_Greek-GDT & 3.23 $\pm$ 0.65
\\ UD\_Marathi-UFAL & 2.81 $\pm$ 0.56 \\ UD\_Turkish-Kenet & 3.08
$\pm$ 0.59 \\ UD\_Frisian\_Dutch-Fame & 2.73 $\pm$ 0.50
\\ UD\_Latin-UDante & 3.66 $\pm$ 0.65 \\ UD\_Icelandic-Modern & 3.35
$\pm$ 0.67 \\ UD\_Norwegian-Bokmaal & 2.93 $\pm$ 0.63
\\ UD\_Makurap-TuDeT & 2.26 $\pm$ 0.44 \\ UD\_Karelian-KKPP & 3.00
$\pm$ 0.52 \\ UD\_Faroese-FarPaHC & 3.63 $\pm$ 0.60 \\ UD\_Livvi-KKPP
& 2.97 $\pm$ 0.54 \\ UD\_Mbya\_Guarani-Dooley & 3.09 $\pm$ 0.40
\\ UD\_Polish-LFG & 2.61 $\pm$ 0.53 \\ UD\_Bambara-CRB & 2.82 $\pm$
0.58 \\ UD\_Russian-GSD & 3.17 $\pm$ 0.55 \\ UD\_Mbya\_Guarani-Thomas
& 3.12 $\pm$ 0.64 \\ UD\_Italian-ISDT & 3.14 $\pm$ 0.62
\\ UD\_Apurina-UFPA & 2.60 $\pm$ 0.55 \\

\hline
\end{tabular}
    
    \end{center}
  \end{minipage}
  \begin{minipage}[t]{.3\textwidth}
    \begin{center}
    \end{center}
  \end{minipage}
  \begin{minipage}[t]{.2\textwidth}
    \begin{center}
    \footnotesize
    \begin{tabular}{lr}
\hline datasets & Binary2 \\ \hline UD\_Turkish\_German-SAGT & 3.13
$\pm$ 0.55 \\ UD\_Icelandic-PUD & 3.28 $\pm$ 0.54
\\ UD\_Chinese-GSDSimp & 3.64 $\pm$ 0.54 \\ UD\_German-LIT & 3.25
$\pm$ 0.56 \\ UD\_Soi-AHA & 2.75 $\pm$ 0.43 \\ UD\_Turkish-Tourism &
2.54 $\pm$ 0.53 \\ UD\_Warlpiri-UFAL & 2.22 $\pm$ 0.41
\\ UD\_Slovak-SNK & 2.71 $\pm$ 0.57 \\ UD\_Czech-PDT & 3.04 $\pm$ 0.66
\\ UD\_North\_Sami-Giella & 2.59 $\pm$ 0.58 \\ UD\_Swiss\_German-UZH &
3.02 $\pm$ 0.45 \\ UD\_Latvian-LVTB & 3.01 $\pm$ 0.62
\\ UD\_Persian-PerDT & 3.30 $\pm$ 0.51 \\ UD\_Komi\_Zyrian-Lattice &
2.90 $\pm$ 0.59 \\ UD\_Hindi-PUD & 3.62 $\pm$ 0.52 \\ UD\_Ukrainian-IU
& 3.02 $\pm$ 0.65 \\ UD\_French-Sequoia & 3.12 $\pm$ 0.77
\\ UD\_Lithuanian-ALKSNIS & 3.10 $\pm$ 0.65 \\ UD\_Vietnamese-VTB &
3.11 $\pm$ 0.55 \\ UD\_Estonian-EDT & 2.98 $\pm$ 0.64
\\ UD\_Indonesian-GSD & 3.32 $\pm$ 0.61 \\ UD\_English-GUM & 2.97
$\pm$ 0.71 \\ UD\_German-HDT & 3.14 $\pm$ 0.59 \\ UD\_Turkish-Penn &
3.00 $\pm$ 0.49 \\ UD\_Russian-SynTagRus & 3.09 $\pm$ 0.59
\\ UD\_English-Pronouns & 2.63 $\pm$ 0.48 \\ UD\_Korean-PUD & 3.50
$\pm$ 0.54 \\ UD\_English-PUD & 3.23 $\pm$ 0.51 \\ UD\_Yoruba-YTB &
3.36 $\pm$ 0.55 \\ UD\_Portuguese-PUD & 3.34 $\pm$ 0.54
\\ UD\_English-EWT & 2.81 $\pm$ 0.77 \\ UD\_Tamil-TTB & 3.47 $\pm$
0.54 \\ UD\_Assyrian-AS & 2.51 $\pm$ 0.57 \\ UD\_Amharic-ATT & 2.96
$\pm$ 0.47 \\ UD\_Romanian-RRT & 3.40 $\pm$ 0.58 \\ UD\_French-FQB &
2.82 $\pm$ 0.44 \\ UD\_Latin-ITTB & 3.11 $\pm$ 0.63 \\ UD\_Tagalog-TRG
& 2.52 $\pm$ 0.50 \\ UD\_Italian-PoSTWITA & 3.18 $\pm$ 0.55
\\ UD\_Turkish-IMST & 2.98 $\pm$ 0.66 \\ UD\_Spanish-PUD & 3.34 $\pm$
0.54 \\ UD\_Irish-TwittIrish & 3.29 $\pm$ 0.62 \\ UD\_Yupik-SLI & 2.92
$\pm$ 0.50 \\ UD\_Breton-KEB & 2.90 $\pm$ 0.57
\\ UD\_Romanian-SiMoNERo & 3.66 $\pm$ 0.56 \\ UD\_Khunsari-AHA & 2.90
$\pm$ 0.30 \\ UD\_Erzya-JR & 2.79 $\pm$ 0.59 \\ UD\_French-Spoken &
2.70 $\pm$ 0.63 \\ UD\_Uyghur-UDT & 3.19 $\pm$ 0.58 \\ UD\_French-PUD
& 3.39 $\pm$ 0.54 \\ UD\_Kiche-IU & 2.58 $\pm$ 0.53
\\ UD\_Sanskrit-UFAL & 2.70 $\pm$ 0.55 \\ UD\_Japanese-PUD & 3.99
$\pm$ 0.52 \\ \hline
\end{tabular}
    
    \end{center}
  \end{minipage}

\end{table} 

\clearpage

\subsection{Averages and standard deviations of lower limits}

\begin{table}[h]
\caption{List of lower limits for each dataset\label{datacount31}}
  \begin{minipage}[t]{.2\textwidth}
    \begin{center}
    \footnotesize
    \begin{tabular}{lr}
\hline datasets & U(n)'s lower limits \\ \hline UD\_Korean-Kaist &
2.75 $\pm$ 0.47 \\ UD\_Faroese-OFT & 2.28 $\pm$ 0.45
\\ UD\_Latin-Perseus & 2.59 $\pm$ 0.56 \\ UD\_Finnish-TDT & 2.58 $\pm$
0.55 \\ UD\_Swedish-LinES & 2.69 $\pm$ 0.56 \\ UD\_Lithuanian-HSE &
2.89 $\pm$ 0.48 \\ UD\_Kaapor-TuDeT & 2.02 $\pm$ 0.13
\\ UD\_Belarusian-HSE & 2.50 $\pm$ 0.56
\\ UD\_South\_Levantine\_Arabic-MADAR & 2.36 $\pm$ 0.48
\\ UD\_Komi\_Zyrian-IKDP & 2.50 $\pm$ 0.53
\\ UD\_Swedish\_Sign\_Language-SSLC & 2.30 $\pm$ 0.59
\\ UD\_Norwegian-NynorskLIA & 2.43 $\pm$ 0.52 \\ UD\_Russian-Taiga &
2.45 $\pm$ 0.55 \\ UD\_Tamil-MWTT & 2.01 $\pm$ 0.09
\\ UD\_Indonesian-CSUI & 3.13 $\pm$ 0.49 \\ UD\_Italian-PUD & 2.98
$\pm$ 0.43 \\ UD\_Swedish-PUD & 2.84 $\pm$ 0.46 \\ UD\_English-ParTUT
& 2.95 $\pm$ 0.50 \\ UD\_Hindi\_English-HIENCS & 2.71 $\pm$ 0.46
\\ UD\_French-ParTUT & 2.97 $\pm$ 0.54 \\ UD\_Gothic-PROIEL & 2.44
$\pm$ 0.55 \\ UD\_Naija-NSC & 2.49 $\pm$ 0.54 \\ UD\_Latin-PROIEL &
2.46 $\pm$ 0.58 \\ UD\_Hungarian-Szeged & 3.00 $\pm$ 0.52
\\ UD\_English-GUMReddit & 2.70 $\pm$ 0.60 \\ UD\_Akkadian-RIAO & 2.54
$\pm$ 0.58 \\ UD\_Chinese-HK & 2.35 $\pm$ 0.51 \\ UD\_Kazakh-KTB &
2.41 $\pm$ 0.51 \\ UD\_German-PUD & 2.89 $\pm$ 0.48 \\ UD\_Serbian-SET
& 2.94 $\pm$ 0.50 \\ UD\_Armenian-ArmTDP & 2.83 $\pm$ 0.62
\\ UD\_Beja-NSC & 2.68 $\pm$ 0.54 \\ UD\_Basque-BDT & 2.59 $\pm$ 0.52
\\ UD\_Croatian-SET & 2.94 $\pm$ 0.51 \\ UD\_Romanian-ArT & 2.54 $\pm$
0.54 \\ UD\_Icelandic-IcePaHC & 2.90 $\pm$ 0.59 \\ UD\_Polish-PDB &
2.70 $\pm$ 0.56 \\ UD\_Afrikaans-AfriBooms & 2.87 $\pm$ 0.50
\\ UD\_Komi\_Permyak-UH & 2.38 $\pm$ 0.51 \\ UD\_Czech-FicTree & 2.51
$\pm$ 0.57 \\ UD\_Old\_East\_Slavic-RNC & 3.07 $\pm$ 0.65
\\ UD\_Turkish-PUD & 2.90 $\pm$ 0.43 \\ UD\_Thai-PUD & 3.07 $\pm$ 0.46
\\ UD\_Spanish-AnCora & 3.14 $\pm$ 0.58 \\ UD\_Persian-Seraji & 2.94
$\pm$ 0.54 \\ UD\_Irish-IDT & 3.04 $\pm$ 0.57 \\ UD\_Finnish-FTB &
2.37 $\pm$ 0.49 \\ UD\_Munduruku-TuDeT & 2.09 $\pm$ 0.29 \\ \hline
\end{tabular}
    
    \end{center}
  \end{minipage}
  \begin{minipage}[t]{.3\textwidth}
    \begin{center}
    \end{center}
  \end{minipage}
  \begin{minipage}[t]{.2\textwidth}
    \begin{center}
    \footnotesize
    \begin{tabular}{lr}
\hline datasets & U(n)'s lower limits \\ \hline UD\_Chinese-PUD & 3.01
$\pm$ 0.47 \\ UD\_Akuntsu-TuDeT & 2.04 $\pm$ 0.20
\\ UD\_Norwegian-Nynorsk & 2.67 $\pm$ 0.56 \\ UD\_Portuguese-Bosque &
2.88 $\pm$ 0.58 \\ UD\_Chukchi-HSE & 2.16 $\pm$ 0.36 \\ UD\_Buryat-BDT
& 2.46 $\pm$ 0.53 \\ UD\_Manx-Cadhan & 2.46 $\pm$ 0.54
\\ UD\_Dutch-Alpino & 2.60 $\pm$ 0.56 \\ UD\_Skolt\_Sami-Giellagas &
2.52 $\pm$ 0.51 \\ UD\_Turkish-FrameNet & 2.22 $\pm$ 0.41
\\ UD\_Japanese-BCCWJ & 2.81 $\pm$ 0.66 \\ UD\_Portuguese-GSD & 3.03
$\pm$ 0.51 \\ UD\_Galician-TreeGal & 2.89 $\pm$ 0.61
\\ UD\_Akkadian-PISANDUB & 2.89 $\pm$ 0.60 \\ UD\_Nayini-AHA & 2.30
$\pm$ 0.46 \\ UD\_French-FTB & 3.08 $\pm$ 0.58 \\ UD\_Korean-GSD &
2.61 $\pm$ 0.57 \\ UD\_Japanese-Modern & 2.74 $\pm$ 0.58
\\ UD\_Wolof-WTB & 2.85 $\pm$ 0.53 \\ UD\_Japanese-GSD & 2.97 $\pm$
0.54 \\ UD\_Classical\_Chinese-Kyoto & 2.07 $\pm$ 0.42
\\ UD\_French-GSD & 3.00 $\pm$ 0.48 \\ UD\_Slovenian-SSJ & 2.75 $\pm$
0.52 \\ UD\_Hebrew-HTB & 3.12 $\pm$ 0.59 \\ UD\_Kangri-KDTB & 2.26
$\pm$ 0.44 \\ UD\_Finnish-OOD & 2.30 $\pm$ 0.55 \\ UD\_Arabic-PUD &
2.98 $\pm$ 0.45 \\ UD\_Low\_Saxon-LSDC & 3.08 $\pm$ 0.55
\\ UD\_Spanish-GSD & 3.06 $\pm$ 0.51 \\ UD\_Old\_East\_Slavic-TOROT &
2.39 $\pm$ 0.51 \\ UD\_Welsh-CCG & 2.83 $\pm$ 0.58 \\ UD\_Moksha-JR &
2.36 $\pm$ 0.48 \\ UD\_Danish-DDT & 2.69 $\pm$ 0.57
\\ UD\_Catalan-AnCora & 3.21 $\pm$ 0.56 \\ UD\_Chinese-CFL & 2.67
$\pm$ 0.59 \\ UD\_Russian-PUD & 2.91 $\pm$ 0.46
\\ UD\_Ancient\_Greek-PROIEL & 2.55 $\pm$ 0.58
\\ UD\_Ancient\_Greek-Perseus & 2.65 $\pm$ 0.56 \\ UD\_Czech-PUD &
2.86 $\pm$ 0.46 \\ UD\_Hindi-HDTB & 2.95 $\pm$ 0.41
\\ UD\_English-LinES & 2.76 $\pm$ 0.55 \\ UD\_Bulgarian-BTB & 2.64
$\pm$ 0.55 \\ UD\_Galician-CTG & 3.11 $\pm$ 0.37 \\ UD\_Urdu-UDTB &
3.06 $\pm$ 0.45 \\ UD\_Indonesian-PUD & 2.94 $\pm$ 0.44
\\ UD\_Turkish-GB & 2.11 $\pm$ 0.35 \\ UD\_Coptic-Scriptorium & 3.00
$\pm$ 0.51 \\ UD\_Old\_French-SRCMF & 2.40 $\pm$ 0.52 \\ \hline
\end{tabular}
    
    \end{center}
  \end{minipage}

\end{table}

\begin{table}[h]
\caption{List of lower limits for each dataset\label{datacount32}}
  \begin{minipage}[t]{.2\textwidth}
    \begin{center}
    \footnotesize
    \begin{tabular}{lr}
\hline datasets & lower limits \\ \hline

UD\_Polish-PUD & 2.88 $\pm$ 0.47 \\ UD\_Italian-TWITTIRO & 2.97 $\pm$
0.31 \\ UD\_Italian-Valico & 2.74 $\pm$ 0.53 \\ UD\_Estonian-EWT &
2.44 $\pm$ 0.61 \\ UD\_Turkish-BOUN & 2.57 $\pm$ 0.56
\\ UD\_Upper\_Sorbian-UFAL & 2.77 $\pm$ 0.55 \\ UD\_Tagalog-Ugnayan &
2.68 $\pm$ 0.47 \\ UD\_Bhojpuri-BHTB & 2.75 $\pm$ 0.52
\\ UD\_German-GSD & 2.78 $\pm$ 0.52 \\ UD\_Sanskrit-Vedic & 2.23 $\pm$
0.44 \\ UD\_Maltese-MUDT & 2.84 $\pm$ 0.60 \\ UD\_Arabic-NYUAD & 3.26
$\pm$ 0.65 \\ UD\_Finnish-PUD & 2.76 $\pm$ 0.48 \\ UD\_Guajajara-TuDeT
& 2.27 $\pm$ 0.44 \\ UD\_Chinese-GSD & 3.12 $\pm$ 0.49
\\ UD\_Tupinamba-TuDeT & 2.23 $\pm$ 0.48 \\ UD\_Slovenian-SST & 2.13
$\pm$ 0.74 \\ UD\_Scottish\_Gaelic-ARCOSG & 2.79 $\pm$ 0.71
\\ UD\_Dutch-LassySmall & 2.44 $\pm$ 0.67 \\ UD\_Italian-ParTUT & 2.98
$\pm$ 0.51 \\ UD\_English-ESL & 2.80 $\pm$ 0.50
\\ UD\_Western\_Armenian-ArmTDP & 2.83 $\pm$ 0.61 \\ UD\_Kurmanji-MG &
2.71 $\pm$ 0.48 \\ UD\_Cantonese-HK & 2.50 $\pm$ 0.59
\\ UD\_Swedish-Talbanken & 2.66 $\pm$ 0.58
\\ UD\_Old\_Church\_Slavonic-PROIEL & 2.38 $\pm$ 0.53
\\ UD\_Telugu-MTG & 2.02 $\pm$ 0.15 \\ UD\_Latin-LLCT & 2.87 $\pm$
0.76 \\ UD\_Old\_Turkish-Tonqq & 2.61 $\pm$ 0.59 \\ UD\_Albanian-TSA &
2.77 $\pm$ 0.42 \\ UD\_Czech-CLTT & 2.88 $\pm$ 0.94
\\ UD\_Romanian-Nonstandard & 2.90 $\pm$ 0.49 \\ UD\_Arabic-PADT &
3.31 $\pm$ 0.70 \\ UD\_Italian-VIT & 2.97 $\pm$ 0.59 \\ UD\_Czech-CAC
& 2.85 $\pm$ 0.56 \\ UD\_Greek-GDT & 2.93 $\pm$ 0.58
\\ UD\_Marathi-UFAL & 2.25 $\pm$ 0.43 \\ UD\_Turkish-Kenet & 2.50
$\pm$ 0.51 \\ UD\_Frisian\_Dutch-Fame & 2.37 $\pm$ 0.48
\\ UD\_Latin-UDante & 3.20 $\pm$ 0.60 \\ UD\_Icelandic-Modern & 2.90
$\pm$ 0.56 \\ UD\_Norwegian-Bokmaal & 2.61 $\pm$ 0.55
\\ UD\_Makurap-TuDeT & 2.03 $\pm$ 0.18 \\ UD\_Karelian-KKPP & 2.62
$\pm$ 0.53 \\ UD\_Faroese-FarPaHC & 3.01 $\pm$ 0.48 \\ UD\_Livvi-KKPP
& 2.61 $\pm$ 0.52 \\ UD\_Mbya\_Guarani-Dooley & 2.53 $\pm$ 0.51
\\ UD\_Polish-LFG & 2.22 $\pm$ 0.41 \\ UD\_Bambara-CRB & 2.51 $\pm$
0.52 \\ UD\_Russian-GSD & 2.87 $\pm$ 0.52 \\ UD\_Mbya\_Guarani-Thomas
& 2.58 $\pm$ 0.55 \\ UD\_Italian-ISDT & 2.81 $\pm$ 0.57
\\ UD\_Apurina-UFPA & 2.20 $\pm$ 0.40 \\ \hline
\end{tabular}
    
    \end{center}
  \end{minipage}
  \begin{minipage}[t]{.3\textwidth}
    \begin{center}
    \end{center}
  \end{minipage}
  \begin{minipage}[t]{.2\textwidth}
    \begin{center}
    \footnotesize
    \begin{tabular}{lr}
\hline datasets & lower limits \\ \hline
UD\_Turkish\_German-SAGT & 2.68 $\pm$ 0.52 \\ UD\_Icelandic-PUD & 2.87
$\pm$ 0.46 \\ UD\_Chinese-GSDSimp & 3.12 $\pm$ 0.49 \\ UD\_German-LIT
& 2.84 $\pm$ 0.53 \\ UD\_Soi-AHA & 2.25 $\pm$ 0.43
\\ UD\_Turkish-Tourism & 2.06 $\pm$ 0.23 \\ UD\_Warlpiri-UFAL & 2.02
$\pm$ 0.13 \\ UD\_Slovak-SNK & 2.40 $\pm$ 0.50 \\ UD\_Czech-PDT & 2.73
$\pm$ 0.57 \\ UD\_North\_Sami-Giella & 2.25 $\pm$ 0.44
\\ UD\_Swiss\_German-UZH & 2.58 $\pm$ 0.51 \\ UD\_Latvian-LVTB & 2.71
$\pm$ 0.57 \\ UD\_Persian-PerDT & 2.82 $\pm$ 0.51
\\ UD\_Komi\_Zyrian-Lattice & 2.53 $\pm$ 0.55 \\ UD\_Hindi-PUD & 3.07
$\pm$ 0.45 \\ UD\_Ukrainian-IU & 2.73 $\pm$ 0.60 \\ UD\_French-Sequoia
& 2.81 $\pm$ 0.66 \\ UD\_Lithuanian-ALKSNIS & 2.83 $\pm$ 0.61
\\ UD\_Vietnamese-VTB & 2.71 $\pm$ 0.47 \\ UD\_Estonian-EDT & 2.59
$\pm$ 0.56 \\ UD\_Indonesian-GSD & 2.92 $\pm$ 0.55 \\ UD\_English-GUM
& 2.71 $\pm$ 0.61 \\ UD\_German-HDT & 2.73 $\pm$ 0.53
\\ UD\_Turkish-Penn & 2.47 $\pm$ 0.51 \\ UD\_Russian-SynTagRus & 2.79
$\pm$ 0.55 \\ UD\_English-Pronouns & 2.12 $\pm$ 0.33 \\ UD\_Korean-PUD
& 2.90 $\pm$ 0.43 \\ UD\_English-PUD & 2.93 $\pm$ 0.44
\\ UD\_Yoruba-YTB & 2.92 $\pm$ 0.42 \\ UD\_Portuguese-PUD & 2.99 $\pm$
0.42 \\ UD\_English-EWT & 2.56 $\pm$ 0.66 \\ UD\_Tamil-TTB & 2.71
$\pm$ 0.52 \\ UD\_Assyrian-AS & 2.23 $\pm$ 0.42 \\ UD\_Amharic-ATT &
2.26 $\pm$ 0.44 \\ UD\_Romanian-RRT & 3.03 $\pm$ 0.47
\\ UD\_French-FQB & 2.40 $\pm$ 0.49 \\ UD\_Latin-ITTB & 2.75 $\pm$
0.57 \\ UD\_Tagalog-TRG & 2.33 $\pm$ 0.47 \\ UD\_Italian-PoSTWITA &
2.86 $\pm$ 0.43 \\ UD\_Turkish-IMST & 2.44 $\pm$ 0.54
\\ UD\_Spanish-PUD & 3.00 $\pm$ 0.43 \\ UD\_Irish-TwittIrish & 2.84
$\pm$ 0.46 \\ UD\_Yupik-SLI & 2.39 $\pm$ 0.49 \\ UD\_Breton-KEB & 2.55
$\pm$ 0.53 \\ UD\_Romanian-SiMoNERo & 3.22 $\pm$ 0.51
\\ UD\_Khunsari-AHA & 2.20 $\pm$ 0.40 \\ UD\_Erzya-JR & 2.45 $\pm$
0.52 \\ UD\_French-Spoken & 2.47 $\pm$ 0.55 \\ UD\_Uyghur-UDT & 2.64
$\pm$ 0.52 \\ UD\_French-PUD & 3.05 $\pm$ 0.43 \\ UD\_Kiche-IU & 2.25
$\pm$ 0.44 \\ UD\_Sanskrit-UFAL & 2.27 $\pm$ 0.44 \\ UD\_Japanese-PUD
& 3.20 $\pm$ 0.45 \\ \hline
\end{tabular}
    
    \end{center}
  \end{minipage}

\end{table} 

\clearpage

\subsection{Averages and standard deviations of upper limits \label{datacountup}}

\begin{table}[h]
\caption{List of upper limits for each dataset \label{datacount41}}
  \begin{minipage}[t]{.2\textwidth}
    \begin{center}
    \footnotesize
    \begin{tabular}{lr}
\hline datasets & Binary2 \\ \hline UD\_Korean-Kaist & 3.52$\pm$0.57
\\ UD\_Faroese-OFT & 2.98$\pm$0.56 \\ UD\_Latin-Perseus & 3.40$\pm$0.70
\\ UD\_Finnish-TDT & 3.34$\pm$0.73 \\ UD\_Swedish-LinES & 3.57$\pm$0.75
\\ UD\_Lithuanian-HSE & 3.85$\pm$0.61 \\ UD\_Kaapor-TuDeT & 2.30$\pm$0.46
\\ UD\_Belarusian-HSE & 3.25$\pm$0.80
\\ UD\_South\_Levantine\_Arabic-MADAR & 3.06$\pm$0.47
\\ UD\_Komi\_Zyrian-IKDP & 3.17$\pm$0.69
\\ UD\_Swedish\_Sign\_Language-SSLC & 2.86$\pm$0.76
\\ UD\_Norwegian-NynorskLIA & 2.97$\pm$0.84 \\ UD\_Russian-Taiga &
3.14$\pm$0.78 \\ UD\_Tamil-MWTT & 2.46$\pm$0.50 \\ UD\_Indonesian-CSUI &
4.17$\pm$0.60 \\ UD\_Italian-PUD & 4.03$\pm$0.57 \\ UD\_Swedish-PUD &
3.85$\pm$0.55 \\ UD\_English-ParTUT & 3.93$\pm$0.65
\\ UD\_Hindi\_English-HIENCS & 3.54$\pm$0.52 \\ UD\_French-ParTUT &
4.03$\pm$0.69 \\ UD\_Gothic-PROIEL & 3.13$\pm$0.76 \\ UD\_Naija-NSC &
3.21$\pm$0.77 \\ UD\_Latin-PROIEL & 3.15$\pm$0.80 \\ UD\_Hungarian-Szeged &
3.99$\pm$0.66 \\ UD\_English-GUMReddit & 3.52$\pm$0.84 \\ UD\_Akkadian-RIAO
& 3.32$\pm$0.72 \\ UD\_Chinese-HK & 2.94$\pm$0.76 \\ UD\_Kazakh-KTB &
3.18$\pm$0.62 \\ UD\_German-PUD & 3.94$\pm$0.56 \\ UD\_Serbian-SET &
3.96$\pm$0.63 \\ UD\_Armenian-ArmTDP & 3.69$\pm$0.81 \\ UD\_Beja-NSC &
3.61$\pm$0.70 \\ UD\_Basque-BDT & 3.42$\pm$0.63 \\ UD\_Croatian-SET &
3.94$\pm$0.65 \\ UD\_Romanian-ArT & 3.32$\pm$0.58 \\ UD\_Icelandic-IcePaHC &
3.86$\pm$0.74 \\ UD\_Polish-PDB & 3.59$\pm$0.68 \\ UD\_Afrikaans-AfriBooms &
3.95$\pm$0.58 \\ UD\_Komi\_Permyak-UH & 3.17$\pm$0.56 \\ UD\_Czech-FicTree &
3.23$\pm$0.80 \\ UD\_Old\_East\_Slavic-RNC & 4.02$\pm$0.90
\\ UD\_Turkish-PUD & 3.82$\pm$0.53 \\ UD\_Thai-PUD & 4.07$\pm$0.57
\\ UD\_Spanish-AnCora & 4.22$\pm$0.74 \\ UD\_Persian-Seraji & 4.03$\pm$0.73
\\ UD\_Irish-IDT & 4.02$\pm$0.76 \\ UD\_Finnish-FTB & 3.03$\pm$0.66
\\ UD\_Munduruku-TuDeT & 2.62$\pm$0.52 \\ \hline
\end{tabular}
    
    \end{center}
  \end{minipage}
  \begin{minipage}[t]{.3\textwidth}
    \begin{center}
    \end{center}
  \end{minipage}
  \begin{minipage}[t]{.2\textwidth}
    \begin{center}
    \footnotesize
    \begin{tabular}{lr}
\hline datasets & U(n)'s upper limits \\ \hline UD\_Chinese-PUD &
3.99$\pm$0.58 \\ UD\_Akuntsu-TuDeT & 2.37$\pm$0.50 \\ UD\_Norwegian-Nynorsk
& 3.52$\pm$0.80 \\ UD\_Portuguese-Bosque & 3.87$\pm$0.82 \\ UD\_Chukchi-HSE
& 2.62$\pm$0.60 \\ UD\_Buryat-BDT & 3.28$\pm$0.66 \\ UD\_Manx-Cadhan &
3.12$\pm$0.52 \\ UD\_Dutch-Alpino & 3.48$\pm$0.71
\\ UD\_Skolt\_Sami-Giellagas & 3.15$\pm$0.64 \\ UD\_Turkish-FrameNet &
2.99$\pm$0.48 \\ UD\_Japanese-BCCWJ & 3.67$\pm$0.95 \\ UD\_Portuguese-GSD &
4.10$\pm$0.66 \\ UD\_Galician-TreeGal & 3.89$\pm$0.81
\\ UD\_Akkadian-PISANDUB & 3.75$\pm$0.83 \\ UD\_Nayini-AHA & 3.10$\pm$0.30
\\ UD\_French-FTB & 4.12$\pm$0.76 \\ UD\_Korean-GSD & 3.35$\pm$0.76
\\ UD\_Japanese-Modern & 3.65$\pm$0.83 \\ UD\_Wolof-WTB & 3.78$\pm$0.68
\\ UD\_Japanese-GSD & 3.96$\pm$0.71 \\ UD\_Classical\_Chinese-Kyoto &
2.59$\pm$0.65 \\ UD\_French-GSD & 4.04$\pm$0.64 \\ UD\_Slovenian-SSJ &
3.61$\pm$0.70 \\ UD\_Hebrew-HTB & 4.18$\pm$0.70 \\ UD\_Kangri-KDTB &
3.03$\pm$0.51 \\ UD\_Finnish-OOD & 2.88$\pm$0.82 \\ UD\_Arabic-PUD &
3.97$\pm$0.57 \\ UD\_Low\_Saxon-LSDC & 4.11$\pm$0.61 \\ UD\_Spanish-GSD &
4.15$\pm$0.64 \\ UD\_Old\_East\_Slavic-TOROT & 3.05$\pm$0.70
\\ UD\_Welsh-CCG & 3.74$\pm$0.71 \\ UD\_Moksha-JR & 3.11$\pm$0.53
\\ UD\_Danish-DDT & 3.53$\pm$0.80 \\ UD\_Catalan-AnCora & 4.28$\pm$0.68
\\ UD\_Chinese-CFL & 3.54$\pm$0.72 \\ UD\_Russian-PUD & 3.89$\pm$0.55
\\ UD\_Ancient\_Greek-PROIEL & 3.32$\pm$0.80
\\ UD\_Ancient\_Greek-Perseus & 3.47$\pm$0.70 \\ UD\_Czech-PUD &
3.81$\pm$0.56 \\ UD\_Hindi-HDTB & 3.89$\pm$0.57 \\ UD\_English-LinES &
3.63$\pm$0.74 \\ UD\_Bulgarian-BTB & 3.46$\pm$0.71 \\ UD\_Galician-CTG &
4.33$\pm$0.50 \\ UD\_Urdu-UDTB & 4.08$\pm$0.62 \\ UD\_Indonesian-PUD &
3.92$\pm$0.56 \\ UD\_Turkish-GB & 2.75$\pm$0.57 \\ UD\_Coptic-Scriptorium &
4.01$\pm$0.68 \\ UD\_Old\_French-SRCMF & 3.09$\pm$0.69 \\ \hline
\end{tabular}
    
    \end{center}
  \end{minipage}

\end{table} 

\begin{table}[h]
\caption{List of upper limits for each dataset \label{datacount42}}
  \begin{minipage}[t]{.2\textwidth}
    \begin{center}
    \footnotesize
    \begin{tabular}{lr}
\hline datasets & U(n)'s upper limits \\ \hline UD\_Polish-PUD &
3.85$\pm$0.55 \\ UD\_Italian-TWITTIRO & 3.92$\pm$0.47 \\ UD\_Italian-Valico
& 3.59$\pm$0.71 \\ UD\_Estonian-EWT & 3.13$\pm$0.86 \\ UD\_Turkish-BOUN &
3.32$\pm$0.75 \\ UD\_Upper\_Sorbian-UFAL & 3.72$\pm$0.66
\\ UD\_Tagalog-Ugnayan & 3.56$\pm$0.56 \\ UD\_Bhojpuri-BHTB & 3.64$\pm$0.68
\\ UD\_German-GSD & 3.78$\pm$0.63 \\ UD\_Sanskrit-Vedic & 2.78$\pm$0.66
\\ UD\_Maltese-MUDT & 3.77$\pm$0.84 \\ UD\_Arabic-NYUAD & 4.25$\pm$0.82
\\ UD\_Finnish-PUD & 3.65$\pm$0.58 \\ UD\_Guajajara-TuDeT & 2.85$\pm$0.47
\\ UD\_Chinese-GSD & 4.09$\pm$0.58 \\ UD\_Tupinamba-TuDeT & 2.80$\pm$0.69
\\ UD\_Slovenian-SST & 2.61$\pm$1.10 \\ UD\_Scottish\_Gaelic-ARCOSG &
3.56$\pm$0.98 \\ UD\_Dutch-LassySmall & 3.15$\pm$1.02 \\ UD\_Italian-ParTUT
& 4.02$\pm$0.66 \\ UD\_English-ESL & 3.71$\pm$0.61
\\ UD\_Western\_Armenian-ArmTDP & 3.71$\pm$0.81 \\ UD\_Kurmanji-MG &
3.55$\pm$0.57 \\ UD\_Cantonese-HK & 3.17$\pm$0.87 \\ UD\_Swedish-Talbanken &
3.54$\pm$0.78 \\ UD\_Old\_Church\_Slavonic-PROIEL & 3.00$\pm$0.77
\\ UD\_Telugu-MTG & 2.51$\pm$0.52 \\ UD\_Latin-LLCT & 3.89$\pm$0.95
\\ UD\_Old\_Turkish-Tonqq & 3.39$\pm$0.59 \\ UD\_Albanian-TSA &
3.68$\pm$0.50 \\ UD\_Czech-CLTT & 3.75$\pm$1.34 \\ UD\_Romanian-Nonstandard
& 3.85$\pm$0.68 \\ UD\_Arabic-PADT & 4.24$\pm$0.90 \\ UD\_Italian-VIT &
3.96$\pm$0.79 \\ UD\_Czech-CAC & 3.79$\pm$0.70 \\ UD\_Greek-GDT & 3.96$\pm$0.79
\\ UD\_Marathi-UFAL & 2.95$\pm$0.61 \\ UD\_Turkish-Kenet & 3.23$\pm$0.66
\\ UD\_Frisian\_Dutch-Fame & 3.05$\pm$0.57 \\ UD\_Latin-UDante &
4.27$\pm$0.72 \\ UD\_Icelandic-Modern & 3.83$\pm$0.73
\\ UD\_Norwegian-Bokmaal & 3.43$\pm$0.77 \\ UD\_Makurap-TuDeT &
2.42$\pm$0.49 \\ UD\_Karelian-KKPP & 3.42$\pm$0.63 \\ UD\_Faroese-FarPaHC &
4.10$\pm$0.64 \\ UD\_Livvi-KKPP & 3.39$\pm$0.63 \\ UD\_Mbya\_Guarani-Dooley
& 3.30$\pm$0.54 \\ UD\_Polish-LFG & 2.89$\pm$0.61 \\ UD\_Bambara-CRB &
3.30$\pm$0.68 \\ UD\_Russian-GSD & 3.86$\pm$0.63 \\ UD\_Mbya\_Guarani-Thomas
& 3.31$\pm$0.73 \\ UD\_Italian-ISDT & 3.77$\pm$0.77 \\ UD\_Apurina-UFPA &
2.89$\pm$0.59 \\ \hline
\end{tabular}
    \end{center}
  \end{minipage}
  \begin{minipage}[t]{.3\textwidth}
    \begin{center}
    \end{center}
  \end{minipage}
  \begin{minipage}[t]{.2\textwidth}
    \begin{center}
    \footnotesize
    \begin{tabular}{lr}
\hline datasets & U(n)'s upper limits \\ \hline
UD\_Turkish\_German-SAGT & 3.52$\pm$0.65 \\ UD\_Icelandic-PUD &
3.85$\pm$0.57 \\ UD\_Chinese-GSDSimp & 4.09$\pm$0.58 \\ UD\_German-LIT &
3.78$\pm$0.68 \\ UD\_Soi-AHA & 2.75$\pm$0.43 \\ UD\_Turkish-Tourism &
2.57$\pm$0.54 \\ UD\_Warlpiri-UFAL & 2.40$\pm$0.49 \\ UD\_Slovak-SNK &
3.10$\pm$0.72 \\ UD\_Czech-PDT & 3.58$\pm$0.81 \\ UD\_North\_Sami-Giella &
2.87$\pm$0.66 \\ UD\_Swiss\_German-UZH & 3.51$\pm$0.62 \\ UD\_Latvian-LVTB &
3.56$\pm$0.76 \\ UD\_Persian-PerDT & 3.74$\pm$0.64
\\ UD\_Komi\_Zyrian-Lattice & 3.25$\pm$0.73 \\ UD\_Hindi-PUD & 4.08$\pm$0.58
\\ UD\_Ukrainian-IU & 3.58$\pm$0.81 \\ UD\_French-Sequoia & 3.70$\pm$0.96
\\ UD\_Lithuanian-ALKSNIS & 3.72$\pm$0.78 \\ UD\_Vietnamese-VTB &
3.55$\pm$0.59 \\ UD\_Estonian-EDT & 3.37$\pm$0.76 \\ UD\_Indonesian-GSD &
3.92$\pm$0.68 \\ UD\_English-GUM & 3.55$\pm$0.90 \\ UD\_German-HDT &
3.71$\pm$0.72 \\ UD\_Turkish-Penn & 3.17$\pm$0.61 \\ UD\_Russian-SynTagRus &
3.69$\pm$0.72 \\ UD\_English-Pronouns & 2.70$\pm$0.49 \\ UD\_Korean-PUD &
3.75$\pm$0.55 \\ UD\_English-PUD & 3.93$\pm$0.55 \\ UD\_Yoruba-YTB &
3.86$\pm$0.54 \\ UD\_Portuguese-PUD & 4.08$\pm$0.57 \\ UD\_English-EWT &
3.29$\pm$0.96 \\ UD\_Tamil-TTB & 3.57$\pm$0.56 \\ UD\_Assyrian-AS &
2.75$\pm$0.76 \\ UD\_Amharic-ATT & 2.99$\pm$0.50 \\ UD\_Romanian-RRT &
4.01$\pm$0.59 \\ UD\_French-FQB & 3.23$\pm$0.49 \\ UD\_Latin-ITTB &
3.58$\pm$0.78 \\ UD\_Tagalog-TRG & 2.86$\pm$0.43 \\ UD\_Italian-PoSTWITA &
3.73$\pm$0.61 \\ UD\_Turkish-IMST & 3.13$\pm$0.75 \\ UD\_Spanish-PUD &
4.09$\pm$0.56 \\ UD\_Irish-TwittIrish & 3.68$\pm$0.63 \\ UD\_Yupik-SLI &
3.08$\pm$0.55 \\ UD\_Breton-KEB & 3.21$\pm$0.69 \\ UD\_Romanian-SiMoNERo &
4.32$\pm$0.59 \\ UD\_Khunsari-AHA & 3.00$\pm$0.45 \\ UD\_Erzya-JR &
3.12$\pm$0.71 \\ UD\_French-Spoken & 3.17$\pm$0.78 \\ UD\_Uyghur-UDT &
3.40$\pm$0.64 \\ UD\_French-PUD & 4.11$\pm$0.57 \\ UD\_Kiche-IU & 3.01$\pm$0.54
\\ UD\_Sanskrit-UFAL & 2.90$\pm$0.65 \\ UD\_Japanese-PUD & 4.28$\pm$0.55
\\ \hline
\end{tabular}
    
    \end{center}
  \end{minipage}
\end{table}

\newpage

\section{Sentences with smallest and largest Strahler numbers}
\label{sec:extremes}

Here are some examples of sentences with the smallest and largest Strahler numbers.

\tabref{strahler1} lists sentences with a Strahler number of 1, all of
which are one word. UD\_English-EWT contains emails, so it includes
many titles, salutations, and signoffs. In addition, it includes
strings that are difficult to separate into words, such as
URLs. UD\_Italian-PoSTWITA is a corpus of tweets, including one-word
tweets. UD\_French-Sequoia is the result of automatic conversion from
another treebank, and it includes titles of paragraphs and chapters.

\begin{table}[h]
\centering
\caption{Examples of sentences with a Strahler number of 1.}
\label{strahler1}
\begin{tabular}{cc}
\hline {Dataset} & Sentence \\ \hline UD\_English-EWT &
Traci\\ UD\_English-EWT & Cheers \\ UD\_English-EWT &
Thanks\\ UD\_English-EWT &
************************************************\\ UD\_English-EWT &
Outlook.jpg\\ UD\_English-EWT & Retired\\ UD\_English-EWT &
VINGAS\\ UD\_Italian-PoSTWITA & polemica\\ UD\_Italian-PoSTWITA &
Chiss\`a\\ UD\_Italian-PoSTWITA & Buongiorno\\ UD\_French-Sequoia&
Espoir\\ UD\_French-Sequoia& NOTICE\\ UD\_French-Sequoia&
R\'ef\'erences\\ UD\_French-Sequoia& \'Epilogue\\
\end{tabular}
\end{table}

The following is an example of a sentence with a Strahler number of
7. It is a legal text in Czech on notes on the consolidating
accounting unit. Sentences with such large Strahler numbers are very
scarce.
\begin{quote}
(1) Konsoliduj\'ic\'i \'u\v{c}etn\'i jednotka uvede v p\v{r}\'iloze v
konsolidovan\'e \'u\v{c}etn\'i z\'av\v{e}rce  a) zp\r{u}sob
konsolidace podle {\S}\_63\_odst.\_1 a pou\v{z}it\'e metody
konsolidace podle  {\S}\_63\_odst.\_4, b) obchodn\'i firmu a
s\'idlo konsolidovan\'ych \'u\v{c}etn\'ich jednotek  zahrnut\'ych
do konsolida\v{c}n\'iho celku; pod\'il na vlastn\'im kapit\'alu v
t\v{e}chto \'u\v{c}etn\'ich  jednotk\'ach zahrnut\'ych do
konsolida\v{c}n\'iho celku dr\v{z}en\'y jin\'ymi \'u\v{c}etn \'imi
jednotkami  ne\v{z} konsoliduj\'ic\'i \'u\v{c}etn\'i jednotkou
nebo osobami jednaj\'ic\'imi vlastn\'im jm\'enem, ale na 
\'u\v{c}et t\v{e}chto \'u\v{c}etn\'ich jednotek; d\'ale uvede
d\r{u}vody, na z\'aklad\v{e} kter\'ych se stala  ovl\'adaj\'ic\'i
osobou, c) obchodn\'i firmu a s\'idlo konsolidovan\'ych
\'u\v{c}etn\'ich jednotek nezahrnut\'ych  do konsolida\v{c}n\'iho
celku podle {\S}\_62\_odst.\_6\_a\_{\S}\_22a\_odst. \\
\_3\_z\'akona,
v\v{c}etn\v{e} d\r{u}vod\r{u}  jejich nezahrnut\'i s uveden\'im
pod\'ilu na vlastn\'im kapit\'alu v t\v{e}chto \'u\v{c}etn\'ich
jednotk\'ach  dr\v{z}en\'eho jin\'ymi osobami ne\v{z}
konsoliduj\'ic\'i \'u\v{c}etn\'i jednotkou, d) obchodn\'i firmu a
s\'idlo  \'u\v{c}etn\'ich jednotek p\v{r}idru\v{z}en\'ych, kter\'e
jsou zahrnuty do konsolidovan\'e \'u\v{c}etn\'i  z\'av\v{e}rky;
pod\'il na vlastn\'im kapit\'alu v t\v{e}chto \'u\v{c}etn\'ich
jednotk\'ach p\v{r}idru\v{z}en\'ych, kter\'y dr\v{z}\'i
\'u\v{c}etn\'i jednotky zahrnut\'e do konsolidace nebo osoby
jednaj\'ic\'i vlastn\'im jm\'enem,  ale na \'u\v{c}et t\v{e}chto
\'u\v{c}etn\'ich jednotek, e) obchodn\'i firmu a s\'idlo
\'u\v{c}etn\'ich jednotek  p\v{r}idru\v{z}en\'ych, kter\'e nejsou
zahrnuty do konsolidovan\'e \'u\v{c}etn\'i z\'av\v{e}rky podle 
{\S}\_62\_odst.\_8, v\v{c}etn\v{e} uveden\'i d\r{u}vodu pro
nezahrnut\'i, f) obchodn\'i firmu a s\'idlo \'u\v{c}etn\'ich 
jednotek pod spole\v{c}n\'ym vlivem zahrnut\'ych do konsolidovan\'e
\'u\v{c}etn\'i z\'av\v{e}rky; pod\'il na  vlastn\'im kapit\'alu v
t\v{e}chto \'u\v{c}etn\'ich jednotk\'ach pod spole\v{c}n\'ym vlivem,
kter\'y dr\v{z}\'i  \'u\v{c}etn\'i jednotky zahrnut\'e do
konsolidace nebo osoby jednaj\'ic\'i vlastn\'im jm\'enem, ale na
\'u\v{c}et  \'u\v{c}etn\'ich jednotek; d\'ale uvede d\r{u}vody, na
z\'aklad\v{e} kter\'ych je vykon\'av\'an spole\v{c}n\'y vliv,  g)
obchodn\'i t\v{e}chto firmu a s\'idlo \'u\v{c}etn\'ich jednotek,
kter\'e nejsou uvedeny pod p\'ismeny  b)\_a\v{z}\_f), v nich\v{z}
maj\'i \'u\v{c}etn\'i jednotky samy nebo prost\v{r}ednictv\'im osoby
jednaj\'ic\'i vlastn\'im  jm\'enem na jej\'i \'u\v{c}et pod\'il na
vlastn\'im kapit\'alu men\v{s}\'i ne\v{z} 20 \%; uvede se v\'y\v{s}e
pod\'ilu na  vlastn\'im kapit\'alu, v\v{c}etn\v{e} celkov\'e
v\'y\v{s}e vlastn\'iho kapit\'alu, v\'y\v{s}e v\'ysledku
hospoda\v{r}en\'i za  posledn\'i \'u\v{c}etn\'i obdob\'i
t\v{e}chto \'u\v{c}etn\'ich jednotek; tato informace nemus\'i b\'yt
uvedena, nejsou-li  tyto \'u\v{c}etn\'i jednotky v\'yznamn\'e z
hlediska pod\'an\'i v\v{e}rn\'eho a poctiv\'eho obrazu
p\v{r}edm\v{e}tu  \'u\v{c}etnictv\'i a finan\v{c}n\'i situace v
konsolidovan\'e \'u\v{c}etn\'i z\'av\v{e}rce, informace o vlastn\'im
kapit\'alu  a o v\'ysledku hospoda\v{r}en\'i se rovn\v{e}\v{z}
neuv\'ad\v{e}j\'i, nejsou-li zve\v{r}ejn\v{e}ny a je-li pod\'il
konsoliduj\'ic\'i  \'u\v{c}etn\'i jednotky na vlastn\'im
kapit\'alu p\v{r}\'imo nebo prost\v{r}ednictv\'im jin\'ych
\'u\v{c}etn\'ich jednotek  men\v{s}\'i ne\v{z} 50 \%, h) informace
o pou\v{z}it\'ych \'u\v{c}etn\'ich metod\'ach a z\'asad\'ach, o
zm\v{e}n\'ach  zp\r{u}sob\r{u} oce\v{n}ov\'an\'i, postup\r{u}
\'u\v{c}tov\'an\'i, uspo\v{r}\'ad\'an\'i polo\v{z}ek konsolidovan\'e
\'u\v{c}etn\'i z\'av\v{e}rky  a obsahov\'eho vymezen\'i
polo\v{z}ek oproti p\v{r}edch\'azej\'ic\'imu \'u\v{c}etn\'imu
obdob\'i, s uveden\'im  d\r{u}vod\r{u} t\v{e}chto zm\v{e}n; u
polo\v{z}ek uveden\'ych v konsolidovan\'e \'u\v{c}etn\'i
z\'av\v{e}rce, kter\'e jsou  nebo p\r{u}vodn\v{e} byly
vyj\'ad\v{r}eny v ciz\'i m\v{e}n\v{e}, se uvedou informace o
zp\r{u}sobu jejich p\v{r}epo\v{c}tu  na m\v{e}nu, v n\'i\v{z} byla
sestavena konsolidovan\'a \'u\v{c}etn\'i z\'av\v{e}rka, i)
vysv\v{e}tlen\'i polo\v{z}ek ,,Kladn\'y  konsolida\v{c}n\'i
rozd\'il''a ,,Z\'aporn\'y konsolida\v{c}n\'i rozd\'il'', metody
jejich stanoven\'i a jak\'ekoli  po\v{c}et v\'yznamn\'e zm\v{e}ny
oproti p\v{r}edch\'azej\'ic\'imu \'u\v{c}etn\'imu obdob\'i, j)
pr\r{u}m\v{e}rn\'y p\v{r}epo\v{c}ten\'y  zam\v{e}stnanc\r{u}
konsolida\v{c}n\'iho celku b\v{e}hem \'u\v{c}etn\'iho obdob\'i, za
kter\'e se sestavuje konsolidovan\'a  \'u\v{c}etn\'i
z\'av\v{e}rka, roz\v{c}len\v{e}n\'ych podle kategori\'i;
samostatn\v{e} se uvede pr\r{u}m\v{e}rn\'y p\v{r}epo\v{c}ten\'y 
po\v{c}et zam\v{e}stnanc\r{u} v pr\r{u}b\v{e}hu \'u\v{c}etn\'iho
obdob\'i u \'u\v{c}etn\'ich jednotek pod spole\v{c}n\'ym vlivem.
\end{quote}

\end{document}